%% file: main.tex
\documentclass[10pt,journal,compsoc]{IEEEtran}
\ifCLASSOPTIONcompsoc
  \usepackage[nocompress]{cite}
\else
  \usepackage{cite}
\fi
\usepackage{amsmath,amsfonts}
\usepackage{algorithmic}
\usepackage{algorithm}
\usepackage{array}
\usepackage[caption=false,font=normalsize,labelfont=sf,textfont=sf]{subfig}
\usepackage{textcomp}
\usepackage{stfloats}
\usepackage{url}
\usepackage{verbatim}
\usepackage{graphicx}
\usepackage{boldline,multirow}
\usepackage{booktabs}
\usepackage{diagbox}

\usepackage{soul}
\usepackage{enumitem}
\usepackage{pdfpages}
\usepackage{flushend}
\usepackage{hhline}
\usepackage[english]{babel}
\usepackage{amsthm}
\usepackage{arydshln}

\newtheorem*{lemma*}{Lemma}
\newtheorem{lemma}{Lemma}[section]
\theoremstyle{definition}
\newtheorem{definition}{Definition}

\newcommand{\indep}{\perp \!\!\! \perp}
\definecolor{stdcolor}{gray}{0.85}
\newcommand{\sgb}[1]{\colorbox{stdcolor}{\scalebox{0.75}{#1}}}

\begin{document}
\newcommand{\lsx}[1]{{\bf\color{cyan}[{\sc Lsx:} #1]}}
\newcommand{\HY}[1]{{\bf\color{red}[{\sc HY:} #1]}}
\newcommand{\CP}[1]{{\bf\color{red}[{\sc CP:} #1]}}

\newcommand{\model}{StableRule}

\title{Rule Learning for Knowledge Graph Reasoning under Agnostic Distribution Shift}

\author{Shixuan Liu*, Yue He*, Yunfei Wang, Hao Zou, Haoxiang Cheng, Wenjing Yang,\\ Peng Cui,~\IEEEmembership{Senior Member,~IEEE}, Zhong Liu

\IEEEcompsocitemizethanks{
\IEEEcompsocthanksitem Shixuan Liu is with the College of Computer Science and Technology, National University of Defense Technology, Hunan, China. This research was conducted during his doctoral studies at the Laboratory for Big Data and Decision, National University of Defense Technology. E-mail: szftandy@hotmail.com
\IEEEcompsocthanksitem Yue He is with the School of Information, Renmin University of China, Beijing, China. E-mail: hy865865@gmail.com
\IEEEcompsocthanksitem Yunfei Wang is with the National Key Laboratory of Information Systems Engineering, National University of Defense Technology, Hunan, China. E-mail: wangyunfei@nudt.edu.cn
\IEEEcompsocthanksitem Hao Zou and Peng Cui are with the Department of Computer Science and Technology, Tsinghua University, Beijing, China. E-mail: ahio@163.com, cuip@tsinghua.edu.cn
\IEEEcompsocthanksitem Haoxiang Cheng and Zhong Liu are with the Laboratory for Big Data and Decision, National University of Defense Technology, Hunan, China. E-mail: hx\_cheng@nudt.edu.cn, liuzhong@nudt.edu.cn
\IEEEcompsocthanksitem Wenjing Yang is with the Department of Intelligent Data Science, College of Computer Science and Technology, National University of Defense Technology, Hunan, China. E-mail: wenjing.yang@nudt.edu.cn}
\thanks{*These authors contributed equally.}
\thanks{(Corresponding authors: Wenjing Yang, Peng Cui.)}
\thanks{This work has been submitted to the IEEE for possible publication. Copyright may be transferred without notice, after which this version may no longer be accessible.}
}


\markboth{IEEE T-PAMI Submission}%
{Liu \MakeLowercase{\textit{et al.}}: Rule Learning for Knowledge Graph Reasoning under Agnostic Distribution Shift}



\input{section/abstract}
\maketitle
\input{section/introduction}
\input{section/related_work}
\input{section/problem2}
\input{section/method}
\input{section/experiments}
\input{section/conclusion}

\section*{Acknowledgments}
This work was supported in part by the National Natural Science Foundation of China (NSFC, 62372459, 62206303, 62001495), China. We would like to express our gratitude to Haotian Wang and Renzhe Xu for their valuable advices.


\bibliographystyle{IEEEtran}
\bibliography{reference}

\input{section/biography}

\clearpage
\input{section/appendix}

\end{document}

%% file: section/abstract.tex
\IEEEtitleabstractindextext{
\begin{abstract}
Logical rule learning, a prominent category of knowledge graph (KG) reasoning methods, constitutes a critical research area aimed at learning explicit rules from observed facts to infer missing knowledge.
However, like all KG reasoning methods, rule learning suffers from a critical weakness-its dependence on the I.I.D. assumption. 
This assumption can easily be violated due to selection bias during training or agnostic distribution shifts during testing (e.g., as in query shift scenarios), ultimately undermining model performance and reliability.
To enable robust KG reasoning in wild environments, this study investigates logical rule learning in the presence of agnostic test-time distribution shifts. 
We formally define this challenge as out-of-distribution (OOD) KG reasoning-a previously underexplored problem, and propose the Stable Rule Learning (\model) framework as a solution. \model~is an end-to-end framework that combines feature decorrelation with rule learning network, to enhance OOD generalization in KG reasoning. 
By leveraging feature decorrelation, \model~mitigates the adverse effects of covariate shifts arising in OOD scenarios, improving the robustness of the rule learning network.
Extensive experiments on seven benchmark KGs demonstrate the framework's superior effectiveness and stability across diverse heterogeneous environments, highlighting its practical significance for real-world applications.

\end{abstract}

\begin{IEEEkeywords}
Knowledge Graph Reasoning, Logical Rule Learning, Distribution Shift
\end{IEEEkeywords}
}

%% file: section/introduction.tex
\section{Introduction}
\begin{figure*}[t]
    \centering
    \includegraphics[width=\linewidth]{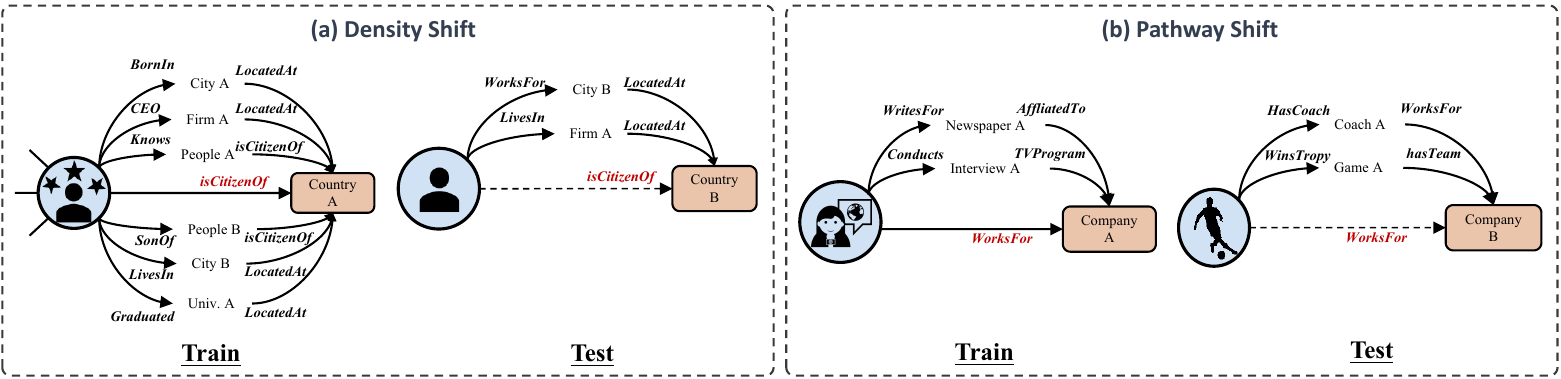}
    \caption{Illustration of KG Reasoning under Query Shift. Solid edges denote observed facts in the training graph, while dashed edges represent target relations to be inferred during testing. (a) Density Shift: Disparity in subgraph density between queried entities. (b) Pathway Shift: compositional reasoning paths scarcely observed during training.}
\label{figure_intro}
\end{figure*} 

\IEEEPARstart{R}{ecent} decades have witnessed the rapid expansion of knowledge graphs (KGs), including widely used resources such as DBPedia~\cite{auer2007dbpedia} and YAGO~\cite{suchanek2007yago}. These KGs offer structured representations of real-world knowledge, proving invaluable across diverse domains, including healthcare~\cite{zhang2020hkgb}, e-commerce~\cite{wang2019kgat}, and social analysis~\cite{abu2021relational}. However, recent research has revealed that some common relationships in existing large KGs are deficient at a rate exceeding $70\%$~\cite{shen2022comprehensive}.
To address this limitation and enable the discovery of new knowledge, KG reasoning remains a critical research focus, aiming to infer missing facts by exploiting observed relational patterns~\cite{chen2020review}.

Currently, three major approaches to KG reasoning exist in the literature. The first is the KG embedding (KGE) method, which encodes entities and relations into a continuous vector space, preserving their semantic similarities as evidenced by structural patterns in the KG~\cite{dai2020survey}. The second paradigm employs graph neural networks (GNNs), utilizing message-passing mechanisms to perform reasoning over the graph structure, thereby facilitating inductive learning of relational dependencies.
The third approach is rule-based reasoning, which derives or learns logical rules by analyzing statistical regularities from sampled or enumerated rule instances (closed paths) within the training graph. Since these rules explicitly represent relational patterns in KGs, they offer interpretable and precise inference capabilities~\cite{yang2017differentiable}. These strands have been demonstrated to be effective under the independent and identically distributed (I.I.D.) hypothesis, where the testing and training data are independently sampled from the same distribution.

In practice, this hypothesis may face significant challenges due to distribution shifts in queries. We could view KG triplets as (query, answer) pairs, where a query comprises a head entity and a relation, and the answer is the tail entity - with the reasoning objective being to generate accurate answers for given queries.
The query shift examined in this paper emerges when the test-time query distribution deviates from the training distribution due to several key factors, including heterogeneous user needs across different demographic groups (e.g., geographical regions or age segments) and temporal variations (e.g., seasonal trends or event-driven spikes in search behavior).

Figure~\ref{figure_intro} presents two representative cases of out-of-distribution (OOD) KG reasoning under query shift. Since most KGE methods are inherently unsuitable for inductive settings with unseen entities, we restrict our analysis to cases where test entities are present in the train graph.
The first example, termed the \textbf{density shift} scenario (Figure~\ref{figure_intro}(a)), demonstrates an imbalance in subgraph density between the queried entities and the correct answer. While the training set primarily contains facts about high-profile public figures, the test set involves nationality queries concerning the general population. 
The disproportionate media coverage of celebrities results in substantially more KG facts about these entities, yielding richer evidence and larger subgraph for nationality inference. 
The second example, termed \textbf{pathway shift}, captures a divergence in reasoning patterns between the queried entities and their answers (see Figure~\ref{figure_intro} (b)). This phenomenon arises due to a selection bias in the training data, wherein the reasoning pathways predominantly reflect the characteristics of journalists, while the test queries primarily pertain to footballers.
Selection bias, a form of sampling bias, occurs when the training data does not adequately represent the test distribution.
In this case, models trained to predict the \textit{WorksFor} relation are predominantly exposed to journalism-related facts (e.g., writing for newspapers, conducting interviews), which may not generalize to footballer-centric test queries, as the underlying pathways in the training and test distributions could differ significantly.

These distribution shifts pose significant challenges to existing KG reasoning methods. Under density shift, KGE models exhibit a representational bias toward information-rich entities (receiving denser training representations), leading to degraded inference performance on sparsely represented entities~\cite{pujara2017sparsity}. Besides, current GNN-based KG reasoning methods, which largely disregard OOD generalization, struggle with size generalization problem, particularly when test graphs exhibit substantially different densities~\cite{yehudai2021local, bevilacqua2021size}. Rule-based methods are similarly vulnerable, as incomplete relation paths in sparse subgraphs undermine inference reliability~\cite{cheng2024logical}.
For pathway shift, KGE methods, biased toward frequent but domain-specific relation patterns for journalists, could fail to generalize to footballer-dominated test data. 
Standard GNNs, which rely on local subgraph aggregation (e.g., paths or motifs), often learn spurious correlations between structural scaffolds and labels rather than generalizable causal patterns~\cite{li2022ood}, leading to poor test-time performance.
Lastly, pathway shift also introduces selection bias in rule instances, favoring journalists over footballers, potentially causing estimation errors in footballer-related rule scores due to rule instances scarcity during training~\cite{qu2020rnnlogic}.

Although the problem of OOD generalization has been extensively studied in domains such as regression~\cite{levine2020offline}, natural language processing~\cite{hendrycks2020pretrained}, and computer vision~\cite{zhang2021deep}, it remains largely unexplored in KG reasoning. 
This gap exists primarily due to the non-Euclidean nature of relational KG data, which requires unique reasoning models with varying representation patterns. Consequently, conventional OOD generalization algorithms, designed to learn $P(Y|X)$ on Euclidean data $\{(X, Y)\}$, cannot be directly applied.




Notably, recent data-driven rule learning methods have established a representation-based framework that models rule plausibility score through conditional distribution $P(r_h|\mathbf{r_b})$, leveraging semantic consistency from observed rule samples $\{(\mathbf{r_b},r_h)\}$~\cite{cheng2022rlogic}.
Through careful analysis informed by OOD generalization research, we identify that covariate shifts in the relation path distribution $P(\mathbf{r_b})$ introduces estimation errors in plausibility score like $P(r_h|\mathbf{r_b})$. Our empirical analysis also confirms the presence of selection bias in $P(\mathbf{r_b})$, which is a typical source of covariate shift. Besides, since path distributions encode fundamental topological properties of KGs~\cite{tishby2022mean}, analysis over $P(\mathbf{r_b})$ helps gauge distribution shifts and may facilitate the characterization of other sources of shifts in KG reasoning. Given above observations, we aim to mitigate covariate shift in $P(\mathbf{r_b})$.

Parallel to such development, stable learning has emerged as a scalable alternative to conventional OOD methods. 
By addressing covariate shift and maintaining broad utility without requiring domain labels, this approach effectively aligns with the above objective.
Motivated by these two complementary advances, we investigate how to learn logical rules from KGs with selection bias, for reasoning on test triples with agnostic distribution shifts. 
We assume that the test queries are sampled from multiple environments, each characterized by a distinct distribution.
Our objective is to ensure that the learned rules generalize effectively across these diverse and potentially unknown environments, thereby enhancing the stability and robustness of KG reasoning performance.

To achieve above objective, we propose Stable Rule Learning (\model), a novel framework comprising two key components: a feature decorrelation regularizer and a rule learning network. 
First, a backtracking-enhanced rule sampler uniformly draws rule instances across all relations, enabling efficient exploration of potential rule bodies. The rule learning network, which parameterize the plausibility score, then processes these rule instances using an encoder to map rule bodies into latent embeddings, followed by a decoder that predicts the rule head from these embeddings.
Concurrently, the feature decorrelation regularizer computes weights for each rule body embedding. These weights mitigate correlations among embedding variables by reweighting the training samples. The weighted embeddings are then integrated into the prediction loss function, promoting the learning of generalizable rule patterns.

Main contributions in this paper are as below:
\begin{itemize}
    \item To the best of our knowledge, we are the first to investigate logical rule learning under OOD KG reasoning scenario, addressing agnostic distribution shifts in KG test sets. 


    \item We propose \model, an end-to-end framework that integrates feature decorrelation with an encoder-decoder rule learning network. By incorporating reweighting mechanisms, our approach effectively eliminates spurious feature correlations.
    
    \item Through extensive experiments on seven benchmark KGs, we demonstrate that \model~achieves competitive performance with SOTA KG reasoning methods, exhibiting favorable effectiveness and generalization capability across diverse environments.

\end{itemize}

The remainder of this paper is organized as follows. Section 2 provides a systematic review of related works. Section 3 formally defines our problem setting and analyses the OOD problem to motivate our framework. Section 4 present the architecture of our proposed framework. Section 5 demonstrates the empirical results through comprehensive experiments on three critical tasks: statistical relation learning, standard KG completion, and inductive link prediction. Section 6 further investigate the properties of \model, including model convergence behavior, rule interpretation, visualization of feature decorrelation effects, ablation studies and efficiency analysis. Finally, Section 7 concludes with key findings and directions for future research.

%% file: section/related_work.tex
\section{Related Work}
\subsection{Knowledge Graph Reasoning} 
In KG reasoning, structure embedding is a well-established approach that map entities and relations onto latent feature spaces considering distance or tensor decomposition.
Translation embedding models defines relation-dependent translation scoring functions to measure the likelihood of triples based on distance metrics.
TransE~\cite{bordes2013transe} is the base translation model that balance effectiveness and efficiency, engendering numerous translation models~\cite{wang2014knowledge,lin2015learning}.
RotatE~\cite{sun2019rotate}, in particular, models each relation as a rotation in the whole complex space, showing impressive representation capacity.
Tensor decomposition methods decompose the high-dimensional fact tensor into multiple lower-dimensional matrices for inferenece, including RESCAL~\cite{nickel2011three}, DistMult~\cite{yang2014embedding} and ComplEx~\cite{trouillon2016complex}.

Given the graph-structured nature of KGs, GNNs are particularly suitable for KG reasoning. RGCN~\cite{schlichtkrull2018modeling} encodes entities via relation-specific neighborhood aggregation and reconstructs facts using a decoder, but its limited variability in entity representations restricts expressiveness. To address this, M-GNN~\cite{wang2019robust} introduced attention, while COMPGCN~\cite{vashishthRGCN} used entity-relation composition operations.  
Recent GNN-based approaches have explored out-of-KG scenarios, primarily addressing inductive reasoning for unseen entities. However, this setting differs fundamentally from conventional OOD generalization problems in traditional machine learning.
GraIL~\cite{teru2020inductive} applies RGCN to local subgraphs for inductive reasoning, while NBFNet~\cite{zhu2021neural}, A*Net~\cite{zhu2023net}, and RED-GNN~\cite{zhang2022knowledge} optimize GNN propagation for efficiency.  

Rule-based method is another branch for reasoning over KGs, known for their interpretability and generalizability to unseen entities. These methods originate from inductive logic programming~\cite{muggleton1994inductive,muggleton1990efficient}, but their scalability remains a challenge due to the NP-hard steps involved~\cite{rocktaschel2017end}. Rule learning methods thus learn rules and weights in a differentiable way. 
Neural-LP~\cite{yang2017differentiable} adopts an attention-based controller to score a rule through differentiable tensor multiplication, but could over-estimate scores for meaningless rules.
DRUM~\cite{sadeghian2019drum} addresses this by using RNNs to sift out potentially incorrect rule bodies.
Since these approach still involve large matrix multiplication, RNNLogic~\cite{qu2020rnnlogic} try to solve scalability issue by partitioning rule generation and rule weight learning, but still struggles with KGs with hundred-scale relations or million-scale entities. 
RLogic~\cite{cheng2022rlogic} thus defines a predicate scoring model trained with sampled rule instances to improve efficiency. 
NCRL~\cite{cheng2022neural} further define the score of a logical rule based on the semantic consistency between rule body and rule head. 

None of above methods take the prevalent distribution shifts between the training and testing data into account, an aspect that this paper investigates for the first time.

\subsection{Out-of-Distribution Generalization}
Domain generalization aims to learn a model using data from a single or multiple interrelated but distinctive source domains, in such a way that the model can generalize well to any OOD target domain.
By assuming the heterogeneity of the training data, these methods employ extra environment labels to learn invariant models capable of generalizing to unseen and shifted testing data. 
Numerous approaches fall under the category of domain alignment, where the primary objective is to reduce the divergence among source domains to learn domain-invariant representations~\cite{li2018learning, li2018domain, dou2019domain, hu2020domain, piratla2020efficient, seo2020learning}.
Some studies expand the available data space by augmenting source domains~\cite{carlucci2019domain, shankar2018generalizing, volpi2018generalizing}. Furthermore, a handful of approaches have benefitted from exploiting regularization through meta-learning and invariant risk minimization~\cite{li2019episodic, arjovsky2019invariant}.

Since the presence of correlations among features can disrupt or potentially undermine model predictions, recent studies have centered their efforts on diminishing such correlations during the training process. 
Some pioneering works theoretically bridge the connections between correlation and model stability under mis-specification, and propose to address such a problem via a sample reweighting scheme~\cite{kuang2020stable, shen2020stable}.
The optimization of the weighted distribution relies on a group of learned sampling weights designed to remove the correlations among covariates.
It has been theoretically established that these models can leverage only those variables that encapsulate invariant correlations for predictions~\cite{xu2021stable}.
Given such property, these methods are suitable for addressing the covariate shift problem in a more pragmatic scenarios where only single-source training data is provided~\cite{zhang2021deep, li2022ood, dou2022decorrelate}. 

%% file: section/problem2.tex
\section{Problem Definition and Analysis}

This section establishes the definitions, formalizes the problem, and presents its analysis. For notational consistency, all key terms and symbols are standardized in Table~\ref{table_notation}.

\input{section/table_notation}

\subsection{Rule Learning Task}



This subsection formally defines the KG reasoning and rule learning task. We adapt certain notations to precisely characterize the OOD rule learning task, which will be introduced in the following subsection.

\begin{definition}[Knowledge Graph Reasoning]  
A KG is formally defined as a structured representation $G = \{E, R, \mathcal{F}_o\}$, where $E$ represents the set of entities, $R$ denotes the set of relations, and $\mathcal{F}_o$ comprises the set of observed facts. Each fact in $\mathcal{F}_o$ is expressed as a triple $(e_h, r, e_t)$, where $e_h, e_t \in E$ and $r \in R$. Alternatively, facts can be interpreted as query-answer pairs $(q, a)$, where the query takes the form $q = (e_h, r, ?)$ and the answer is $a = e_t$.  
The primary objective of KG reasoning is to model the conditional distribution $P(a|G, q)$ for inferring answers to queries.
During testing, KG reasoning models are assessed on a set of unobserved test facts $\mathcal{F}_u=\{(q',a')\}$.
\end{definition}  


Logical rules play a pivotal role in enhancing KG reasoning, as they provide interpretable and logically grounded explanations for predictive tasks. Most rule-learning methods focus on chain-like compositional Horn rules to facilitate scalable reasoning over KGs. These rules are formally defined as follows:

\begin{definition}[Horn Rule]
A compositional Horn rule is comprised of a body of conjunctive relations and a head relation, as follows:
\begin{equation}
S(\mathbf{r_b}, r_h): r_{b_1}(x,z_1) \wedge \cdots \wedge r_{b_l}(z_{l-1},y) \rightarrow r_h(x,y)
\end{equation}
where $S(\mathbf{r_b}, r_h)\in [0, 1]$ represents the rule's plausibility score, and the antecedent $r_{b_1}(x,z_1) \wedge \cdots \wedge r_{b_l}(z_{l-1},y)$ constitutes the \textbf{rule body} (a relation path of length $l$), and the consequent $r_h(x,y)$ denotes the \textbf{rule head} (target relation for reasoning). 
\end{definition}

When variables in such rules are instantiated with concrete entities, we obtain a \textbf{rule instance} (alternatively termed a closed path). For notational convenience, we represent rules as $(\mathbf{r_b}, r_h)$, where $\mathbf{r_b} = [r_{b_1}, \cdots, r_{b_l}]$ compactly encodes the rule body.

\begin{definition}[Logical Rule Learning]
The logical rule learning task aims to learn a plausibility score $S(\mathbf{r_b}, r_h)$ for each
rule $(\mathbf{r_b}, r_h)$ in the rule space $\mathcal{R}$, quantifying its quality. 
\end{definition}

During inference, the top-$k$ highest-scoring rules are selected as the learned rule set. Given these scored rules, we can efficiently compute scores for candidate answers through forward chaining~\cite{salvat1996sound} (see Appendix C).


\subsection{Rule Learning Task in OOD Setting}


Current KG reasoning methods are predominantly evaluated under the assumption that test facts follow the same distribution as training facts, as both sets are typically derived from a single, complete KG via random splitting. This evaluation protocol inherently ensures identical distributions between training and test queries. However, such an assumption may not hold in practical scenarios, where query distributions can shift due to dynamic and diverse user demands, as illustrated in Section 1, while the training data may contain abundant facts about journalists, the actual reasoning task may primarily involve queries about footballers and their affiliated companies. Such discrepancies represent a form of agnostic shift, where query distribution shifts naturally arise in real-world applications. To rigorously characterize this challenge, we introduce the following definition:  

\begin{definition}[Out-of-Distribution Knowledge Graph Reasoning]
The OOD KG reasoning task involves $\{G, \mathcal{F}_o, \mathcal{F}_u\}$. During training, a model learns the conditional distribution $P(a|G, q)$ from $G$ and $\{(q,a)\}\subseteq \mathcal{F}_o$
At test time, it is evaluated on diverging sets of queries $\textbf{q'}_m$ drawn from $M$ heterogeneous test environments, i.e., $\mathcal{F}_u=\{(\textbf{q'}_m, \textbf{a'}_m)\}_{m=1}^M$.
Crucially, for each relation $r\in R$, the distribution of subject entities $P(e_h)$ should vary across test environments to induce meaningful distribution shifts.
\end{definition}

The fundamental challenge of OOD KG reasoning stems from systematic shifts in query distributions. For a certain relation $r\in R$ (e.g., WorksFor), we enforce shifts in the head entity distribution $P(e_h)$ across test subgroups - for instance, creating one group dominated by journalists and another by footballers. We distinguish two principal types of distributional shifts:
\begin{itemize}
    \item \textbf{Natural Shifts}: Subgroups partitioned by observable attributes (e.g., professions or nationalities).  
    \item \textbf{Synthetic Shifts}: Since exhaustively accounting for exogenous factors is infeasible, an alternative is to classify head entities based on their relation paths to answers. 
\end{itemize}


If a KG reasoning model can reliably learn the conditional probability $P(a|G, q)$ despite variations in query distributions $P(q)$, it should exhibit stability and efficacy across all $M$ heterogeneous environments. 
To address this OOD reasoning challenge, the goal of rule learning in OOD setting is to accurately learn the plausibility score for each rule $(\mathbf{r_b}, r_h)$ in the rule space $\mathcal{R}$, via identification and mitigation of the adverse effects of query shifts. This ensures that predictions based on these scores remain both accurate and stable across diverse environments.



\subsection{Problem Analysis}
\label{sec:clarify}

 

\subsubsection{The Impact of Query Shift}

As previously discussed, maintaining stable performance under query shifts is our key objective. However, directly measuring performance fluctuations in response to such shifts presents significant methodological challenges. 
To overcome this, we analyze their effects through the lens of relation path distribution $P(\mathbf{r_b})$, which fundamentally governs fact inference for rule learning methods. 

We first analyse the phenomenon of pathway shift induced by query shift, as shown in Figure 1(b). 
When the training triples contain significantly more journalists than footballers, rule-learning methods will observe a higher frequency of relation paths semantically linked to journalists. For example, when reasoning the relation \textit{WritesFor}, rule bodies such as $\mathbf{r_b}_1$: \textit{WritesFor} $(x,z) \wedge$ \textit{AffiliatedTo} $(z,y)$ and $\mathbf{r_b}_2$: \textit{Conducts} $(x,z) \wedge$ \textit{TVProgram} $(z,y)$ exhibit stronger statistical relevance to journalist head entities. In contrast, $\mathbf{r_b}_3$: \textit{HasCoach} $(x,z) \wedge$ \textit{WorksFor} $(z,y)$ and $\mathbf{r_b}_4$: \textit{WinsTropy} $(x,z) \wedge$ \textit{hasTeam} $(z,y)$ are more strongly associated with footballers. 
As a result, for journalist head entities,  $\mathbf{r_b}_1$ and $\mathbf{r_b}_2$ appear more frequently in closed paths obtained through path enumeration than $\mathbf{r_b}_3$ and $\mathbf{r_b}_4$.
Consequently, in a training graph dominated by journalists, the sampled rule instances yield higher marginal probabilities $P(\mathbf{r_b}=\mathbf{r_b}_1)$ and $P(\mathbf{r_b}=\mathbf{r_b}_2)$ compared to $P(\mathbf{r_b}=\mathbf{r_b}_3)$ and $P(\mathbf{r_b}=\mathbf{r_b}_4)$.  
Reversing the journalist-footballer ratio in the KG would invert this probability distribution. Importantly, footballer-related relations do not disappear from the training graph - while their representation becomes imbalanced, their presence remains essential for learning.

Second, we examine the phenomenon of density shift. As depicted in Figure 1(a), this shift occurs when test queries predominantly concern the public, while the rule samples used for training are primarily derived from celebrities. 
This discrepancy induces a shift in node degree and path sparsity, likely altering the underlying distribution of relation paths $P(\mathbf{r_b})$ with high probability.


These observations reveal that selection bias manifests in relation paths, typically inducing a covariate shift in the distribution of relation paths $P(\mathbf{r_b})$, ultimately impairing model generalization. 
Furthermore, the reasoning performance of rule-based models depends heavily on rule quality, as measured by plausibility scores. Therefore, our objective reduces to correcting misjudgments in these plausibility scores induced by covariate shifts in $P(\mathbf{r_b})$.





\subsubsection{Plausibility Score Selection}

Selecting an appropriate plausibility score is critical for rule learning, yet traditional metrics from associational rule mining~\cite{galarraga2013amie}, such as coverage and confidence, fail to characterize distribution shifts properly. We believe an effective score for OOD rule learning should be grounded in representation learning, computationally scalable, and explicitly incorporate relation paths to address these limitations:

Firstly, representation-based scores are particularly valuable for OOD generalization because they extract domain-invariant features robust to distribution shifts. By disentangling high-level semantic patterns from low-level relational combinations, such representations enhance generalization where raw features may lack discriminative power. 
Besides, scalability is equally essential, especially when query shift happen in large-scale KGs where rule space risk combinatorial explosion. 
In IID settings, the rule space is often constrained to patterns seen during training, making scalability less critical. However, under covariate shift, models may need to explore a broader rule space (e.g., longer or more complex paths).
Lastly, embedding relation paths into the score’s definition enables direct integration with existing OOD algorithms designed to mitigate covariate shift. 

With above insights in mind, we note that recent data-driven approaches could formalize this intuition by constructing plausibility scores from the semantic consistency of rule instances $(\mathbf{r_b}, r_h)$. These scores, expressed as $P(r_h | \mathbf{r_b})$, model the probability of the rule head $r_h$ replacing the rule body $\mathbf{r_b}$ in closed-path samples. This formulation not only supports efficient, high-performance reasoning by incorporating path probabilities $P(\mathbf{r_b})$ but also aligns with the canonical OOD objective $P(Y|X)$.  
Consequently, our challenge further boils down to eliminating estimation errors arising from covariate shifts in relation path distributions $\mathbf{r_b}$ while effectively learning $P(r_h | \mathbf{r_b})$.


\subsubsection{Reducing Covariate Shift}


Learning $P(r_h|\mathbf{r_b})$ becomes challenging when facing distribution shifts in $P(\mathbf{r_b})$. Although balancing query group sizes (e.g., equalizing journalists and footballers) during training could potentially mitigate such shifts, 
it relies on domain expert explicitly revealing the groups to balance, which is labor-intensive and inefficient. In contrast, stable learning, which can balance all the potential groups and consequently reduce the spurious correlation, becomes a satisfactory strategy.

From a machine learning perspective, when estimating $P(Y|X)$ from samples $\{(X, Y)\}_N$, sample selection bias introduce covariate shifts in $P(X)$, creating spurious correlations in the input space~\cite{chen2016robust,sugiyama2013learning}. If representations $Z$ are learned as intermediate features (via $X$ to $Z$ to $Y$), these covariate shifts propagate to $Z$. 
Notably, while feature decorrelation has proven effective for tabular/structured data (where features have fixed semantics), recent work demonstrates that operating on learned representations $Z$ yields superior results for complex modalities like images and graphs~\cite{zhang2021deep,li2022ood}.
We therefore posit that minimizing statistical dependencies in $Z$ offers a principled approach to reducing these errors~\cite{xu2022theoretical}.

Under the background of rule learning, where the objective is to learn $P(r_h|\mathbf{r_b})$, sample bias in $P(\mathbf{r_b})$ can introduce covariate shifts and spurious correlations in the learned representation $P(\textbf{Z})$. By eliminating statistical dependencies within $P(\textbf{Z})$, we could improve the robustness of $P(r_h|\mathbf{r_b})$ estimation under covariate shifts in $P(\mathbf{r_b})$.
Building on this insight, we propose to use reweighting methods to mitigating covariate shifts in the embedding space $P(\mathbf{Z})$. 
This approach aligns with stable learning principles, which enhances robustness under distribution shifts by decorrelating input features (or their representations) through reweighting. 
The reweighting scheme is particularly suitable for our setting because it implicitly handles unknown group structures without manually providing groups for balancing in prior, has been empirically validated for robustness under covariate shifts, and operates flexibly on learned representations to accommodate high-dimensional rule features.

%% file: section/table_notation.tex
\begin{table}[h!]
\renewcommand\arraystretch{1.2}
\centering
\caption{Notation and Definitions.}
\resizebox{\linewidth}{!}{
\begin{tabular}{cp{0.75\linewidth}}
 \toprule[1.5pt]       
\textbf{Notation} & \textbf{Definition} \\ 
 \midrule
$E$ & Set of entities in the knowledge graph. \\
$R$ & Set of relations in the knowledge graph. \\
$G = (E, R, \mathcal{F}_o)$ & Knowledge graph subject to reasoning. \\
$\mathcal{F}_o = \{(q, a)\}$ & Set of observed facts used for model training, where each fact could be considered as a query-answer pair. \\
$\mathcal{F}_u = \{(q', a')\}$ & Set of unobserved facts used for evaluation. \\
$q = (e_h, r, ?)$ & A query to be answered involves predicting the tail entity given a head entity and a relation. \\
$a = (e_t)$ & The answer to a query, representing the tail entity $e_t$. \\
$\mathbf{q'}_m$ & Query set in the $m$-th test environment. \\
$M$ & Total number of distinct test environments. \\
$\mathbf{r_b}$ & Rule body, defined as a conjunction of relations. \\
$r_h$ & Rule head, denoting a target relation to be inferred. \\
$S(\mathbf{r_b}, r_h)$ & Plausibility score assigned to the rule $(\mathbf{r_b}, r_h)$. \\
$N$ & Batch size for rule samples during training. \\
$d$ & Embedding dimensionality for rule bodies. \\
$\mathbf{Z}_{n*}$ & Embedding representation of the $n$-th rule body 
\\
$\mathbf{Z}_{*i}$ &  The random variable in the $i$-th dimension. \\
$\mathbf{W}=\{w_n\}^N_{n=1}$ &
The weights associated with the rule samples, where $w_n$ is the weight for the $n$-th rule sample
\\
 \bottomrule[1.5pt]
\end{tabular}
}
\label{table_notation}
\end{table}


%% file: section/method.tex
\section{Methodology}
\begin{figure*}[t]
    \centering
    \includegraphics[width=\linewidth]{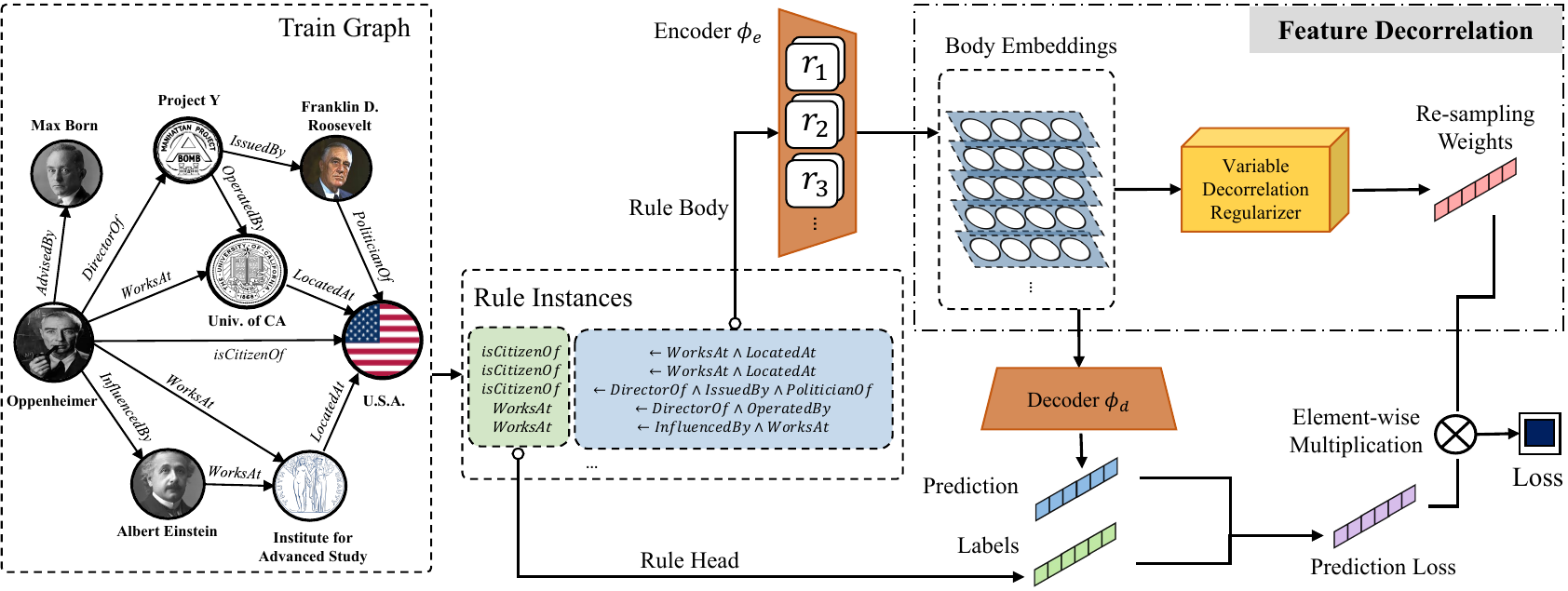}    \caption{Overview of \model. We initially generate a batch of rule instances $\{(\mathbf{r_b}, r_h)\}_N$ from the train graph. The rule bodies are translated into representations with the encoder $\phi_e$.
    Subsequently, the decoder $\phi_d$ learns to classify these embeddings into corresponding rule heads. To eliminate spurious correlations present in $\mathbf{r_b}$, we derive a batch of weights $\textbf{W}$, with feature decorrelation, to appropriately re-weight the training samples. The overall optimization goal is to minimize the weighted loss.}
\label{figure_model}
\end{figure*} 


This section introduces our proposed \model~framework, designed to maintain stable reasoning performance under query shifts. 
The framework's primary objective is to learn a stable predictor for $P(r_h|\mathbf{r_b})$ that remains effective even when covariate shifts occur in $P(\mathbf{r_b})$, thereby supporting reliable KG reasoning under distinct environments.

\subsection{Overview}
As illustrated in Figure \ref{figure_model}, our framework starts with sampling a batch of rule instances $\{(\mathbf{r_b}, r_h)\}_N$ from the training graph, which are subsequently processed by an encoder-decoder framework to parameterize $P(r_h | \mathbf{r_b})$.
The rule encoder, denoted as $\phi_e:\mathcal{R}_b\rightarrow \mathbf{Z}$, maps rule bodies from the input space $\mathcal{R}_b$ to a $d$-dimensional latent representation space $\mathbf{Z}$. Conversely, the decoder $\phi_d:\mathbf{Z} \rightarrow \mathcal{R}_h$ reconstructs rule heads from their latent representations. 
For a batch of $N$ rule bodies $\{\mathbf{r_b}\}_N$, their corresponding representations are collectively denoted as $\mathbf{Z}$.  


As noted in Section~\ref{sec:clarify}, a critical challenge arises from covariate shifts in $\mathbf{r_b}$, which induce covariate shifts in $\mathbf{Z}$.
Specifically, spurious correlations can emerge due to varying dependencies among the covariates of $\mathbf{Z}$ across different environments. To mitigate this bias, we propose a sample re-weighting strategy that enforces statistical independence among the dimensions of $\mathbf{Z}$. This approach enhances estimation robustness by counteracting the adverse effects of covariate shifts. 


Formally, let $\mathbf{Z}_{n*}$ represent the latent embedding of the $n$-th rule body, and $\mathbf{Z}_{*i}$ denote the random variable associated with the $i$-th dimension of $\mathbf{Z}$. The sample weights $\mathbf{W} = \{w_n\}_{n=1}^N$ are constrained such that $\sum_{n=1}^N w_n = N$, where $w_n$ corresponds to the weight assigned to the $n$-th rule instance $(\mathbf{r_b}, r_h)_n$. Through joint optimization of the encoder $\phi_e$, decoder $\phi_d$, and sample weights $\mathbf{W}$, we ensure that the learned representations $\mathbf{Z}$ exhibit minimal statistical dependence across dimensions under the re-weighted distribution. This facilitates the generalization of the composed predictor $\phi_e \circ \phi_d: \mathcal{R}_b \rightarrow \mathcal{R}_h$ to shifted queries. 

Since eliminating dependency is the key ingredient for effective rule learning in OOD scenarios, we prioritize the exposition of our methodology for mitigating statistical dependence (Section~\ref{sec:reweight}) before detailing the parameterization of the rule learning network (Section~\ref{sec:rulelearn}) and the overall training scheme (Section~\ref{sec:trainscheme}).

\subsection{Statistical Independence with Re-weighting}
\label{sec:reweight}
In this subsection, we formalize our approach for achieving statistical independence in the learned representations.
Crucially, the intrinsic dependencies within the body representations, particularly between relevant components\footnote{The relevant components correspond to discriminative features critical for predicting rule heads, which maintain invariant relationships with the rule heads under covariate shifts (e.g., semantically-related or synonymous predicates).} and irrelevant components\footnote{The irrelevant components consist of non-predictive or spurious features, such as noisy relations, redundant semantics without logical dependencies, or coincidental structural patterns (e.g., specific relation path segments) that overfit to training data but lack generalizability.}, easily lead to the estimation error of $P(r_h|\mathbf{r_b})$\cite{arjovsky2019invariant, tu2020empirical}. 

When these components exhibit strong interdependencies, they generate spurious correlations that compromise the model's predictive validity.
Our approach eliminates statistical dependencies across all dimensions of the rule body representations. Formally, we implement this via:

\begin{equation}
\mathbf{Z}_{*i} \indep \mathbf{Z}_{*j}, \forall i, j \in [1, d], i \neq j.
\end{equation}

\noindent From~\cite{bisgaard2006does}, we know that $\mathbf{Z}_{*i}$ and $\mathbf{Z}_{*j}$ are independent if,
\begin{equation}
\mathbb{E}[\mathbf{Z}_{*i}^a\mathbf{Z}_{*j}^b] = \mathbb{E}[\mathbf{Z}_{*i}^a]\mathbb{E}[\mathbf{Z}_{*j}^b], \forall a,b \in \mathbb{N}
\end{equation}
where $\mathbf{Z}_{*i}^a$ represents the element-wise exponentiation of $\mathbf{Z}{i}$ raised to the power $a$. Drawing upon established weighting methodologies in the causal inference literature~\cite{athey2018approximate, fong2018covariate}, we propose to enforce statistical independence between $\mathbf{Z}_{i}$ and $\mathbf{Z}_{*j}$ through sample reweighting using a weight matrix $\mathbf{W}$. 
The optimal weights can be derived by minimizing the following objective function:
\begin{equation}
\label{eqa3}
\sum_{a=1}^{\infty} \sum_{b=1}^{\infty} ||\mathbb{E}[{\mathbf{Z}_{*i}^a}^T \mathbf{\Sigma}_\mathbf{W} \mathbf{Z}_{*j}^b] - \mathbb{E}[{\mathbf{Z}_{*i}^a}^T \mathbf{W}]\mathbb{E}[{\mathbf{Z}_{*j}^b}^T \mathbf{W}]||^2
\end{equation}
where $\mathbf{\Sigma}_\mathbf{W} = diag(w_1, \cdots, w_n)$ is the corresponding diagonal matrix. 
While enforcing exact moment conditions for all variable orders in the objective function (Equation~\eqref{eqa3}) may be impractical, theoretical insights from~\cite{kuang2020stable} demonstrate that reducing first-order moment dependencies can enhance estimation accuracy and model stability. In practice, to balance empirical performance while accounting for potential higher-order correlations in the representation space, we treat the moment order $O_d$ as a hyperparameter in our framework. Consequently, the optimization problem in Equation~\eqref{eqa3} can be effectively implemented by minimizing the following loss function:

\begin{equation}
L_d = \sum_{i=1}^d \sum_{a=1}^{O_d}\sum_{b=1}^{O_d} ||{\mathbf{Z}_{*i}^a}^T\mathbf{\Sigma}_\mathbf{W} \mathbf{Z}_{*\neg i}^b/N - {\mathbf{Z}_{*i}^a}^T \mathbf{W}/N*{\mathbf{Z}_{*\neg i}^b}^T \mathbf{W}/N||^2
\label{eqa4}
\end{equation}
where $\mathbf{Z}{*\neg i}$ denotes the complement of the $i$-th variable in $\mathbf{Z{*}}$, representing all remaining variables when $\mathbf{Z}_{i}$ is excluded. The summand quantifies the dependence penalty between the target variable $\mathbf{Z}{*i}$ and its complement $\mathbf{Z}{*\neg i}$, capturing their residual statistical association.

Here, we prove that $W$ is learnable and admits at least one solution. While prior work~\cite{kuang2020stable} established this result for the special case of first-order moments $(a=b=1)$ in Equation (\ref{eqa4}), we extend this to arbitrary moment orders.

\begin{lemma}
\label{lem}
If the number of covariates d is fixed, as $h_i \rightarrow 0$ for $i=1,\cdots, d$ and $Nh_1 \cdots h_d \rightarrow \infty$, then exists a $W \succeq 0$ such that  $\lim_{n\rightarrow \infty}L_d = 0$ 
with probability 1. In particular, a solution $\textbf{W}$ that satisfies $\lim_{n\rightarrow \infty}L_d = 0$ is $w_n^*=\frac{\prod_{i=1}^d \hat{f}(\textbf{Z}_{ni})}{\hat{f}(\textbf{Z}_{n1},\cdots, \textbf{Z}_{nd})}$, where $\hat{f}(\textbf{Z}_{*i})$ and $\hat{f}(\textbf{Z}_{*1}, \cdots, \textbf{Z}_{*d})$ are the Kernel density estimators.
\end{lemma}

\begin{proof}
From~\cite{hollander2013nonparametric}, we know if $h_i \rightarrow 0$ for $i=1,\cdots, d$ and $N h_1 \cdots h_d \rightarrow \infty$, 
$\hat{f}(z_{ni})= f(z_{ni})+o_p(1)$
and
$\hat{f}(z_{n*})= f(z_{n*})+o_p(1)$

Note that for any $j$, $a$, 
\begin{equation*}
\begin{aligned}
  1/N \sum_{n=1}^N &\mathbf{Z}_{nj}^a w_n = 1/N \sum_{n=1}^N \mathbf{Z}_{nj}^a \frac{\prod_{i=1}^d f(\textbf{Z}_{ni})}{f(\textbf{Z}_{n1},\cdots \textbf{Z}_{nd})} +o_p(1) \\
  &= \mathbb{E}[\mathbf{Z}_{nj}^a \frac{\prod_{i=1}^d  f(\textbf{Z}_{ni})}{f(\textbf{Z}_{n1},\cdots \textbf{Z}_{nd})}] +o_p(1) \\
  &=  \int \cdots \int \mathbf{Z}_{nj}^a \prod_{i=1}^d f(\textbf{Z}_{ni})d\textbf{Z}_{n1}\cdots d\textbf{Z}_{nd} +o_p(1) \\ 
  &=  \int \mathbf{Z}_{nj}^af(z_{nj})d \textbf{Z}_{nj} +o_p(1)
\end{aligned}
\end{equation*}

Similarly, for any $k$, $b$,
\begin{equation*}
  1/N \sum_{n=1}^N \mathbf{Z}_{nk}^b w_n = \int \mathbf{Z}_{nk}^bf(z_{nk})d \textbf{Z}_{nk} +o_p(1)
\end{equation*}

Therefore, for any $j$ and $k$, $j \neq k$ and any $a$ and $b$, 
\begin{equation*}
\begin{aligned}
  1/N \sum_{n=1}^N &\mathbf{Z}_{nj}^a\mathbf{Z}_{nk}^b w_n = 1/N \sum_{n=1}^N \mathbf{Z}_{nj}^a\mathbf{Z}_{nk}^b \frac{\prod_{i=1}^d f(\textbf{Z}_{ni})}{f(\textbf{Z}_{nj},\cdots \textbf{Z}_{nd})} +o_p(1) \\
  &= \mathbb{E}[\mathbf{Z}_{nj}^a\mathbf{Z}_{nk}^b \frac{\prod_{i=1}^d f(\textbf{Z}_{ni})}{f(\textbf{Z}_{n1},\cdots \textbf{Z}_{nd})}] +o_p(1) \\
  &= \int \int \mathbf{Z}_{nj}^a \mathbf{Z}_{nk}^b f(z_{nj})f(z_{nk})d\textbf{Z}_{nj}d\textbf{Z}_{nk} +o_p(1)\\ 
  &= \int \mathbf{Z}_{nj}^a f(z_{nj})d \textbf{Z}_{nj}* \int \mathbf{Z}_{nk}^b f(z_{nk})d \textbf{Z}_{nk} +o_p(1)
\end{aligned}
\end{equation*}

Thus, 
\begin{equation*}
\begin{aligned}
  \lim_{N\rightarrow \infty} L_d &= \lim_{N\rightarrow \infty} \sum_{a=1}^{O_d}\sum_{b=1}^{O_d}  \sum_{j\neq k} (1/N \sum_{n=1}^N \mathbf{Z}_{nj}^a\mathbf{Z}_{nk}^b w_n - \\
  &(1/N \sum_{n=1}^N \mathbf{Z}_{nj}^a w_n)(1/N \sum_{n=1}^N \mathbf{Z}_{nk}^b w_n))^2=0
\end{aligned}
\end{equation*}
\end{proof}

This theoretical result demonstrates that our proposed variable decorrelation regularizer can effectively decorrelate spurious correlations in $\textbf{Z}$ by sample reweighting.

\begin{algorithm}[t]
   \caption{Stable Rule Learning (\model) framework}
   \label{alg:example}
\begin{algorithmic}[1]
   \STATE {\bfseries Input:} knowledge graph $G$ and observed facts $O$.
   \STATE Initialize the parameters in the encoder $\phi_e$ and the decoder $\phi_d$ for the rule learning network.
   \STATE Generate a pool of rule samples with walking strategy described in Section~\ref{subsec:sample}.
    \FOR{$i=1$ to $T_r$}
        \STATE Sample a batch of rule samples $\{(\mathbf{r_b}, r_h)\}_N$
        \STATE Compute body embeddings $\mathbf{Z}$ with $\phi_e({\mathbf{r_b}}_N)$.
        \STATE Randomly initialize re-sampling weights $\mathbf{W}$.
        \FOR{$j=1$ to $T_d$}
        \STATE Optimize $\mathbf{W}$ by minimizing Equation~\eqref{eqa4} by rate $\alpha_d$.
        \ENDFOR
    \STATE Optimize $\phi_e, \phi_d$ in the direction of Equation~\eqref{equ:overall} with learning rate $\alpha_r$. 
    \ENDFOR

   \STATE {\bfseries Output:} network parameters for $\phi_e$ and $\phi_d$.
\end{algorithmic}
\end{algorithm}

\subsection{Rule Learning Network}
\label{sec:rulelearn}

\subsubsection{Neural Network Architecture}

Various neural architectures can be employed to construct the encoder and decoder for parameterizing $P(r_h|\mathbf{r_b})$. A critical aspect of this process involves deriving effective embeddings for the input rule body sequence $\mathbf{r_b} = [r_{b_1}, \cdots, r_{b_l}]$. 
While existing sequence models, such as long short-term memory networks (LSTMs) and Transformers, excel at capturing sequential dependencies, they often fail to model the underlying deductive structure (such as a compositional tree) that governs the derivation of the target relation.
To address this limitation, we present an implementation of the encoder-decoder framework that explicitly accounts for the deductive structure. Our design is motivated by the merits of the deductive nature of rules, as discussed in Appendix D.




The encoder employs a sliding window approach to process relation sequences, where each window of fixed size $s$ at position $k$ is defined as the subsequence $[r_{b_k}, \ldots, r_{b_{k+s}}]$. These windows are encoded into continuous representations using an LSTM parameterized by $\theta$: $\Omega_k=\mbox{LSTM}_\theta([r_{b_{k}}, \ldots, r_{b_{k+s}}])$. 
The resulting window embeddings $\Omega$ are subsequently projected through a fully connected layer to compute a probability distribution $Q$ over the potential logical compositions formed by the windows. 
This transformation is formulated as $Q = \sigma(W_f \Omega + b_f)$, 
where $\sigma(\cdot)$ represents the softmax activation function, $W_f \in \mathbb{R}^{1 \times d}$ denotes the weight matrix, and $b_f \in \mathbb{R}$ is the bias term, with $d$ being the window embeddings dimensionality. The window with the maximal probability $Q_k$ is selected for combination.  
Finally, an attention mechanism is employed to compute a weighted combination of all head relation representations, thereby replacing the original window representation $\Omega_k$ by $\Omega_k'$:

\begin{equation}
\Omega_k' = Attention(\Omega_k) \cdot H W_2=\sigma (\frac{\Omega_k W_1(HW_2)^T}{\sqrt{d}}) H W_2
\end{equation}
where, $W_1,W_2 \in \mathbb{R}^{d\times d}$ are trainable weights, 
and $H$ represent the relation embedding table.
Through recursive composition merging over the rule body until convergence to a single window representation, the encoder generates the final body representation $\mathbf{Z}_{n*}$. The decoder module share the attention mechanism to compute the head predicate likelihood $P(r_h|Z)$ for each body representation $\mathbf{Z}_{n*}$ as $Attention(\mathbf{Z}_{n*})$.


\subsubsection{Backtracking-enhanced Rule Sampling} 
\label{subsec:sample}


Rather than exhaustively enumerating all closed paths in the KG, we adopt a path sampling strategy to generate a subset of closed paths as rule instances for model training. To enable efficient sampling, we propose a backtracking-enhanced random walk procedure inspired by principles from stochastic processes~\cite{spitzer2013principles}.  
Specifically, the process begins by uniformly sampling a batch of facts $\{(e_h, r, e_t)\}$ for each relation $r \in R$. Then, starting from a source entity $e_h$, we perform a random walk of fixed length $l$, where each subsequent entity $e_{i}$ is selected based on the transition probability $P(e_i|e_{i-1}) = \frac{1}{|N(e_{i-1})|}$, given that $(e_i, r, e_{i-1}) \in \mathcal{F}_o$. 

For the backtracking mechanism, at each step $i$, we check whether a direct relation exists between the starting entity $e_h$ and the current entity $e_i$. If such a relation is found, we record the path as a valid rule instance, where the head relation $r_h$ corresponds to the connecting edge and the body $\mathbf{r_b}$ represents the traversed path. 
Additionally, we incorporate negative sampling by considering non-closed paths. If no relation links $e_h$ and $e_i$ at any point during the walk, we generate a rule sample with $r_h$ labeled as "Neg". This approach ensures an effective exploration of negative rule instances while maintaining computational efficiency.



\subsection{Objective Function}
\label{sec:trainscheme}

Our framework jointly optimizes sample weights $\mathbf{W}$ and the encoder-decoder parameters through an alternating minimization procedure. Unifying the objectives of feature decorrelation and rule learning, we formulate the composite optimization problem as:

\begin{equation}
\label{equ:overall}
\hat{\phi_e}, \hat{\phi_d} = \underset{\phi_e, \phi_d}{\arg \min} \sum^n_{i=1}w_i L(\phi_d(\phi_e(\mathbf{r_b^i})), r_h^i)
\end{equation}
where $L(\cdot, \cdot)$ denotes the cross-entropy loss function. This choice of loss function reflects the rule learning objective, which seeks to maximize the likelihood of the observed target relation in closing the sampled path.



%% file: section/experiments.tex
\section{Experiments}

\begin{figure*}[h!]
\subfloat{\includegraphics[width = .166\linewidth]{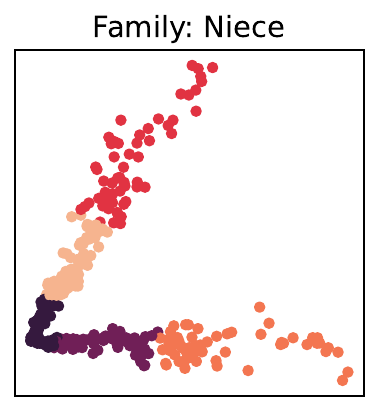}} 
\subfloat{\includegraphics[width = .166\linewidth]{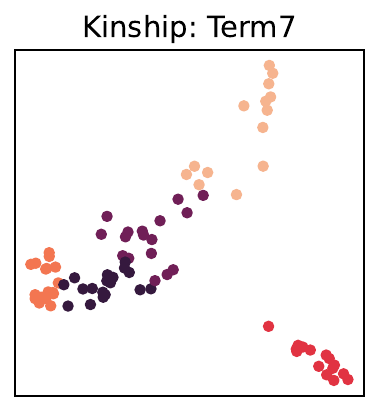}}
\subfloat{\includegraphics[width = .166\linewidth]{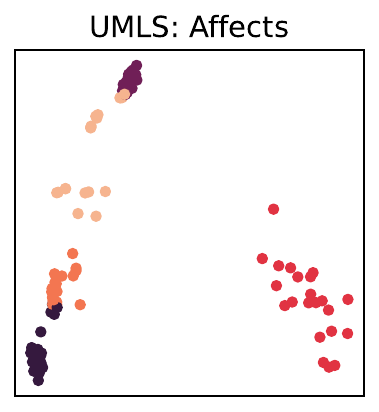}}
\subfloat{\includegraphics[width = .166\linewidth]{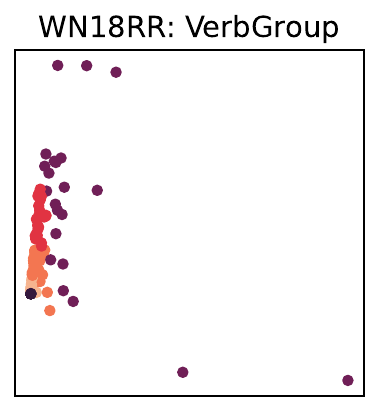}} 
\subfloat{\includegraphics[width = .166\linewidth]{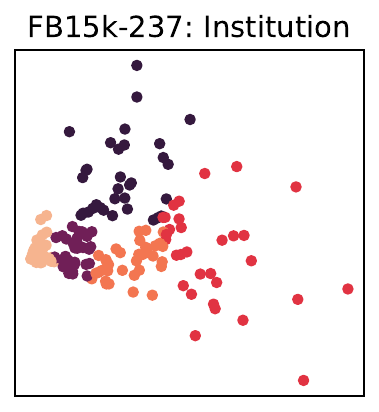}} 
\subfloat{\includegraphics[width = .166\linewidth]{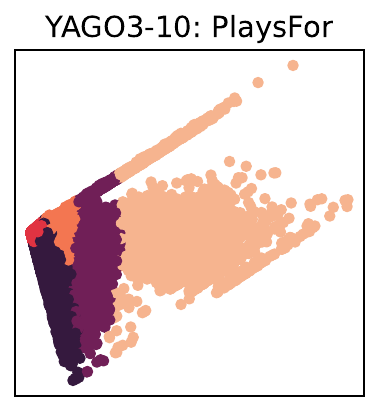}} 
\caption{Visualization of the partitioning of test set over one relation. 
We can see the queries are divided into distinct groups.
}
\label{fig_cluster}
\end{figure*}

To validate the efficacy of our framework in KG reasoning under agnostic distribution shift, we conduct three key experiments across diverse subgroups: statistical relation learning, standard KG completion (KGC), and inductive link prediction. 
In statistical relation learning, we assess natural distribution shifts by focusing on reasoning over a single relation.
For KGC, we construct synthetic test sets via clustering to ensure query shifts among different test splits.
Inductive link prediction follows the same configuration as KGC but operates under a strictly entity-disjoint setting between training and test graphs.
The rationale for these experimental designs is elaborated below:

\begin{itemize}
    \item In statistical relation learning, we focus on reasoning the \textit{Author} relation in the DBLP dataset, which captures paper authorship, to investigate real-world distribution shifts. This setup serves two purposes: (1) examining how an exogenous factor (country) induces natural query shifts and leads to imbalanced model performance across test subgroups, and (2) enabling an in-depth analysis of relation path statistics. Variations in body coverage across subgroups reveal why certain models underperform in specific subgroup.  

    
    \item KGC serves as a standard benchmark for evaluating KG reasoning methods, requiring models to comprehend all relations and answer to queries involving both $(e_h, r, ?)$ and $(?, r, e_t)$ after a single training phase. To induce systematic query shifts, we generate synthetic test sets through clustering-based partitioning.


    \item While KGC typically operates in a transductive setting, where distribution shifts may still arise, inductive link prediction enforces entity disjointness between training and test graphs. This setting could better reflects real-world OOD scenarios and often induces more pronounced shifts, as evidenced by observable performance degradation.  
    
\end{itemize}



\begin{table}[h!]
\renewcommand\arraystretch{1.0}
\centering
\caption{Statistics of adopted benchmark KGs.}
\resizebox{0.9\linewidth}{!}{%
\begin{tabular}{cccc}
 \toprule[1.5pt]
  Dataset & \# Entities & \# Relations & \# Facts 
  \\ \hline
  Family &3,007 &12  &28,356\\
  Kinship & 104 & 25 & 9,587  \\
  UMLS & 135 & 46 & 5,960\\
WN18RR & 40,943 & 11 & 93,003 \\
FB15K-237 & 14,541 & 237 & 310,116 \\
YAGO3-10 & 123,182 & 37 & 1,089,040 \\
DBLP & 559,396 & 4 & 2,381,868 \\
    \bottomrule[1.5pt]
\end{tabular}
}
\label{table_dataset}
\end{table}

\begin{figure*}[htb]
    \centering
    \subfloat{\includegraphics[width=\linewidth]{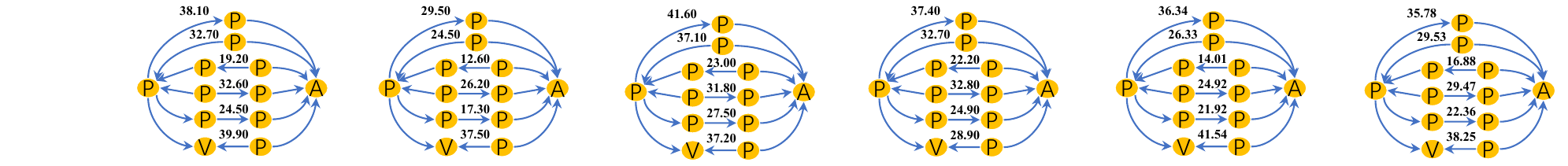}} \\
    \subfloat{\includegraphics[width=\linewidth]{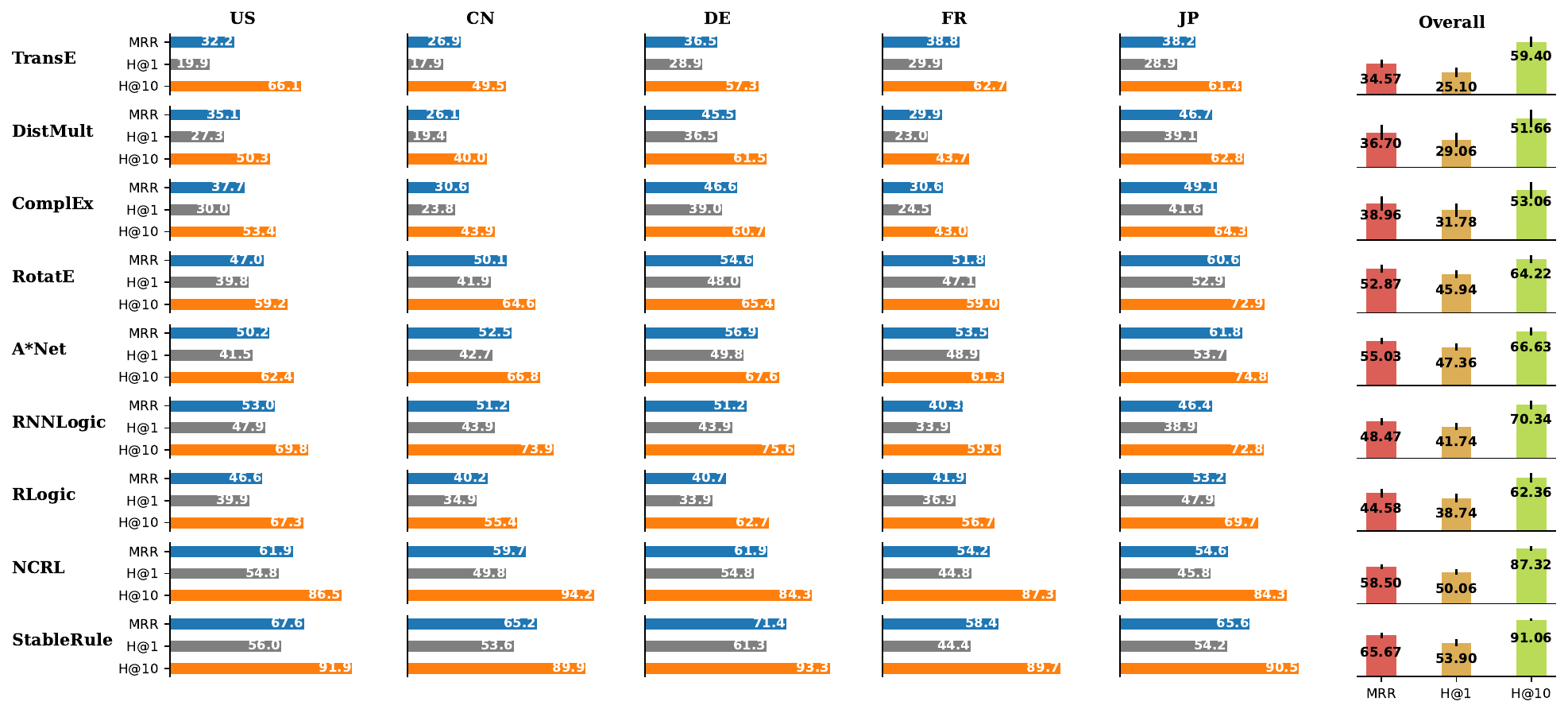}    }
    \caption{Results on DBLP over subgroups. The above part illustrates the variations in body coverage (which gauges the frequency of each body, shown in $\%$) of selected rules of high confidence, for each respective subgroup and the average population. For example, the 38.10 value (top left) in the US group suggests that 38.10\% of Paper-Author pairs are linked through a path instance of Cites (Paper, Paper)$\wedge$Author (Paper, Author). Such variation across groups could help explain performance differences. The below part, from left to right, delivers a performance evaluation for each subgroup, followed by the overall performance.}
    
\label{figure_dblp}
\end{figure*} 

\subsection{Experimental Settings}
\noindent \textbf{Datasets.} To evaluate the effectiveness and robustness of our framework, we conduct experiments on seven widely-used benchmark KGs. The selected datasets encompass diverse domains and scales: Family~\cite{hinton1986learning}, Kinship~\cite{kok2007statistical}, UMLS~\cite{kok2007statistical} for smaller-scale evaluation; WN18RR~\cite{dettmers2018convolutional}, FB15K-237~\cite{toutanova2015observed}, YAGO3-10~\cite{suchanek2007yago} and DBLP~\cite{tang2008arnetminer} for medium to large-scale assessment. Statistics for these datasets are presented in Table~\ref{table_dataset}, while a detailed description of their characteristics is provided in Appendix A.


\noindent \textbf{Utilizing Natural Shift.} In statistical relational learning, we introduce a real-world exogenous factor (i.e., country affiliation) to partition queries within the DBLP dataset. This approach enables us to examine whether such natural query shifts induce imbalanced model performance across subgroups. Specifically, we evaluate models on their ability to infer authorship for groups of papers where authors are affiliated with different countries. Author country labels are derived using the GeoText package~\cite{chen2013geotext}, based on the most frequently occurring affiliation for each individual. We restrict our analysis to papers authored by individuals from the top five countries in terms of publication volume (abbreviated as US, CN, DE, FR, and JP).


\noindent \textbf{Creating Synthetic Shift.} We utilize the first six datasets to conduct KGC task under synthetic query shifts. To construct test sets that emulate query shifts across different testing environments, we partition the original test set into five approximately equally sized subgroups. This division is achieved through a controlled clustering procedure applied to the path-describing vector, which quantifies the path counts between each query-answer pair for all possible rule bodies.
To ensure a balanced distribution of queries across subgroups for each relation, the clustering operation is performed on a per-relation basis. This approach guarantees distinct query shifts among the test subgroups. The resulting clusters, corresponding to individual relations from each of the six datasets, are visually depicted in Figure~\ref{fig_cluster}.


\noindent \textbf{Baselines.} We compare our method against 12 competitive approaches, spanning KGE, GNN-based, and rule-learning methods. A comparison of baselines is provided in Table~\ref{tab:baseline}, with detailed descriptions included in Appendix B.

\input{section/table_baselines}


\noindent \textbf{Evaluation Metrics.} During testing, we evaluate each method by masking either the head or tail entity of every test triple and requiring the model to predict the missing entity. To ensure a rigorous assessment, we employ a filtered evaluation setting and measure performance using three standard metrics: Hit@1, Hit@10, and Mean Reciprocal Rank (MRR). To enhance statistical reliability, we compute the results for each subgroup as the average over five independent runs. The final KG completion performance is then determined by averaging the results across all five subgroups. The results are presented in the form of mean $\pm$ standard deviation.


\noindent \textbf{Experimental Setup.} Our experiments are conducted on a high-performance computing server equipped with a 64-core CPU, 256 GB of RAM, and eight NVIDIA RTX-3090Ti GPUs, each with 24GB of memory. The proposed \model~framework is implemented using PyTorch~\cite{paszke2019pytorch} and trained in an end-to-end fashion. For optimization, we employ the Adam optimizer~\cite{kingma2014adam}. The embeddings of all relations are initialized uniformly to ensure consistent starting conditions. To ensure a fair comparison with baseline methods, we determine the optimal hyper-parameters for all competing baselines using a comprehensive grid search strategy. Detailed hyper-parameter configurations for our method are provided in Appendix E.

\input{section/table_main}

\input{section/table_no_shift}

\subsection{Statistical Relation Learning}
The DBLP results are presented in Figure~\ref{figure_dblp}. As illustrated in Figure~\ref{figure_dblp} (above), a significant disparity in body coverage exists among several top-ranked rules, confirming the presence of selection bias in $P(\mathbf{r_b})$.
Furthermore, the findings indicate that path patterns can vary substantially, reflecting differences in research collaboration patterns across the selected countries.

Figure~\ref{figure_dblp} (below) demonstrates that \model~outperforms all baseline methods, exhibiting a clear performance gap in both effectiveness and stability across all subgroups.
KGE models struggle to generate suitable embeddings for all entities within this extensive KG, even after an extended training period of 48 hours.
For GNN-based approaches, while A*Net demonstrates better scalability than NBFNet (which fails to scale entirely), its performance remains suboptimal on this particular dataset.
Furthermore, in contrast to \model, other rule-learning methods display heightened sensitivity to natural query shift across subgroups, resulting in reduced and unstable performance.


Notably, each method underperforms in specific subgroups. Our analysis reveals that the worst-case subgroup consistently corresponds to the CN and FR groups. Further examination indicates significant deviations in path patterns for these groups compared to the average population. These findings reinforce our proposition that developing reasoning methods robust to agnostic distribution shifts is critical for their real-world application.

\subsection{Knowledge Graph Completion}
The KGC results are presented in Table~\ref{table_subpop}. Neural-LP, DRUM, and RNNLogic are unable to accommodate the extensive size of the YAGO3-10 dataset, and we exclude them from our reported results.
For OOD scenario evaluation, we emphasize the importance of examining both mean performance metrics and their standard deviations. This dual assessment ensures we maintain overall effectiveness while preventing either: (1) disproportionate impacts on lower-performing groups, or (2) diminishing performance across all groups to reduce variance.
The most stable model achieves superior overall performance while maintaining small variance across groups.

In comparison to KGE approaches, \model~demonstrates superior performance over three established KGE models, namely TransE, DistMult, and ComplEx. Among KGE methods, RotatE emerges as the strongest competitor, maintaining competitive performance across the Family, Kinship, and UMLS datasets. However, this relative advantage decreases with increasing KG scale, since larger KGs could exhibit higher structural heterogeneity and therefore more frequent distribution shifts.
GNN-based approaches, particularly NBFNet and A*Net, demonstrate competitive performance on the WN-18RR dataset, benefiting from their ability to propagate and combine information along relation-conditioned paths. Nevertheless, \model~outperforms these GNN-based approaches across the remaining five datasets.
When compared with other rule learning methods, \model~consistently achieves superior performance with evident performance gains. Our model exhibits relatively lower variance, indicating more consistent performance across different test sets - a characteristic attributable to its robustness against covariate shift. 

Beyond this, we note that KGE methods generally surpass rule-based approaches and typically exhibit smaller standard deviations across diverging subgroups. This advantage stems primarily from their capacity to learn effective entity representations, particularly in smaller-scale KGs. 
The inherent smoothness of similarity metrics derived from embedding proximity naturally yields lower variance compared to rule-based similarity measures, which are inherently more susceptible to KG incompleteness.
However, these strengths are noticeably diminished in inductive scenarios involving unseen entities, which is more pervasive in real-world OOD scenarios. 


Although our method is primarily designed for OOD scenarios, it remains effective in the absence of query shifts. As demonstrated in Table~\ref{table_no_shift}, while the performance gap between our method and baselines is somewhat reduced compared to the OOD setting, our approach still consistently outperforms other rule learning methods. This persistent advantage stems primarily from our decorrelation module's capability to reduce selection bias inherent in the path sampling process.

\input{section/table_inductive}
\begin{figure*}[t]
    \centering    \includegraphics[width=\linewidth]{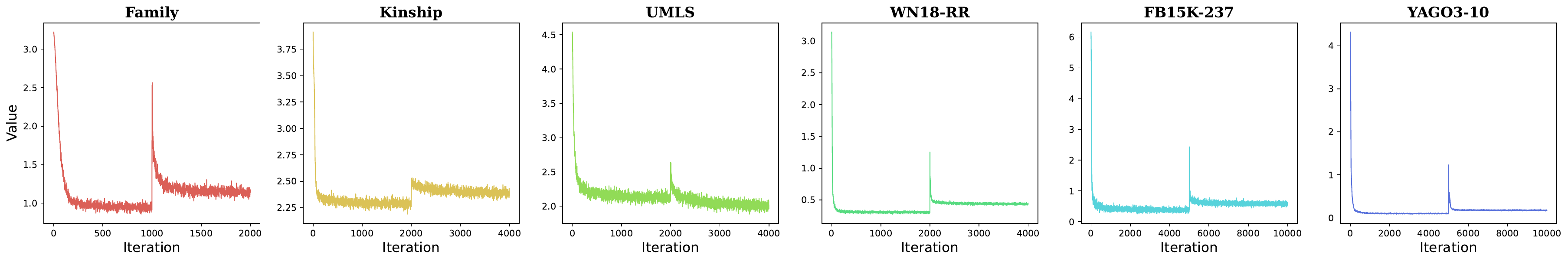}    \caption{Loss curves on six datasets. Each curve is averaged over 3 runs.}
\label{figure_loss}
\end{figure*} 


\input{section/table_rules}

\subsection{Inductive Link Prediction}
It should be emphasized that drawing direct comparisons between rule-based methods and KGE methods based solely on transductive completion tasks is inherently inequitable. Unlike KGE approaches, which lack the capability to reason about unseen entities, rule-based methods demonstrate particular strength in inductive scenarios.
To construct such inductive scenario, we randomly partitioned facts in the KG into training (70\%), validation (10\%), and test (20\%) sets, ensuring complete entity disjointness between splits. The test set was further divided into subgroups following the methodology described in the KGC experiment.


Table~\ref{table_ind} presents the inductive link prediction results. While GNN-based methods show smaller performance declines than in the transductive setting, benefiting from their inherent inductive reasoning capabilities, our method achieves superior performance through its robust learning capacity.
Furthermore, our analysis reveals that although all rule-based methods experience performance degradation in this setting, \model~achieves larger performance advantage over other rule-based approaches compared to the transductive scenario.
This notable difference demonstrates \model's particular strength in inductive reasoning tasks.

%% file: section/table_baselines.tex
\begin{table}[htb]
\centering
\caption{Categories of baseline methods}
\resizebox{0.9\linewidth}{!}{
\begin{tabular}{cccc}
\toprule[1.5pt]
Names & KGE  & GNN   &  Rule Learning                               \\ \hline
\textbf{TransE}~\cite{bordes2013transe}                 &   \checkmark       &      &          \\ 
\textbf{DistMult}~\cite{yang2014embedding}                 &   \checkmark       &    &          \\ 
\textbf{ComplEx}~\cite{trouillon2016complex}              &   \checkmark       &    &          \\ 
\textbf{RotatE}~\cite{sun2019rotate}     &  \checkmark   &        &                  \\ 
\textbf{CompGCN}~\cite{vashishthRGCN}       &       &   \checkmark       &                              \\ 
\textbf{NBFNet}~\cite{zhu2021neural}            &    &   \checkmark       &                              \\ 
\textbf{A* Net}~\cite{zhu2023net}       &   &  \checkmark        &                           \\ 
\textbf{Neural-LP}~\cite{yang2017differentiable}               &   &        &  \checkmark            \\ 
\textbf{DRUM}~\cite{sadeghian2019drum}       &    &     &    \checkmark                   \\ 
\textbf{RNNLogic}~\cite{qu2020rnnlogic}                &     &      &   \checkmark          \\ 
\textbf{RLogic}~\cite{cheng2022rlogic}                &         &     &     \checkmark              \\
\textbf{NCRL}~\cite{cheng2022neural}                 &         &        &   \checkmark       \\ 
\bottomrule[1.5pt]

\end{tabular}
}
\label{tab:baseline}
\end{table}

%% file: section/table_main.tex
\begin{table*}
\renewcommand\arraystretch{1.2}
\caption{Knowledge graph completion results across six datasets, averaged over five sub-populations. Bold and underlined values denote the best-performing results for each dataset among all methods and rule-based methods, respectively.}
\resizebox{\linewidth}{!}{%
\begin{tabular}{cccccccccc}
 \toprule[2pt]
\multirow{2}{*}{Methods} & \multicolumn{3}{c}{Family}              & \multicolumn{3}{c}{Kinship}             & \multicolumn{3}{c}{UMLS}                \\
& MRR (\%)   & HIT@1 (\%) & HIT@10 (\%) & MRR (\%)   & HIT@1 (\%) & HIT@10 (\%) & MRR (\%)   & HIT@1 (\%) & HIT@10 (\%) \\

\hline
TransE    & 78.05\sgb{±2.29} & 61.38\sgb{±3.48}  & 97.39\sgb{±1.52}   & 37.68\sgb{±3.46} & 10.55\sgb{±2.01}  & 85.61\sgb{±5.68}   & 73.86\sgb{±3.34} & 52.04\sgb{±5.67}  & 98.19\sgb{±0.73}   \\
DistMult  & 66.62\sgb{±3.66} & 50.99\sgb{±4.47}  & 96.46\sgb{±0.97}   & 41.56\sgb{±2.55} & 26.48\sgb{±2.49}  & 78.20\sgb{±4.27}   & 57.04\sgb{±4.13} & 45.98\sgb{±4.59}  & 79.11\sgb{±5.50}   \\
ComplEx   & 85.43\sgb{±2.18} & 76.24\sgb{±2.96}  & 97.99\sgb{±1.24}   & 58.76\sgb{±2.32} & 45.82\sgb{±2.46}  & 87.35\sgb{±1.94}   & 72.29\sgb{±3.04} & 57.93\sgb{±3.96}  & 93.92\sgb{±1.90}   \\
RotatE    & \textbf{93.85\sgb{±2.36}} & \textbf{89.99\sgb{±3.25}}  & \textbf{99.08\sgb{±1.25}}   & \textbf{74.50\sgb{±5.15}} & \textbf{58.30\sgb{±5.84}}  & \textbf{96.34\sgb{±3.75}}   & \textbf{80.24\sgb{±2.08}} & \textbf{74.89\sgb{±2.99}}  & \textbf{97.04\sgb{±0.83}}   \\

\hdashline
CompGCN    & 75.35\sgb{±2.90} &58.83\sgb{±2.20} &88.83\sgb{±2.84} &40.30\sgb{±2.63} &26.94\sgb{±2.04}&75.83\sgb{±3.08}  &62.33\sgb{±2.23} &50.85\sgb{±3.84}&72.76\sgb{±1.89}  \\
NBFNet   & 84.16\sgb{±2.00} & 76.03\sgb{±2.34}  & 97.71\sgb{±0.38} & 49.54\sgb{±2.78} & 34.43\sgb{±2.74}  & 81.07\sgb{±2.84}   & 70.74\sgb{±2.43} & 64.19\sgb{±3.13}  & 82.48\sgb{±1.46}     \\
A*Net   & 82.35\sgb{±2.69} &72.37\sgb{±3.21} &96.67\sgb{±1.42} &44.36\sgb{±2.79} &30.42\sgb{±1.95} &77.36\sgb{±2.44}   &66.37\sgb{±2.04} &53.63\sgb{±5.51} &  78.37\sgb{±1.47} \\

\hdashline
Neural-LP & 86.12\sgb{±1.75} & 79.28\sgb{±3.01}  & 97.53\sgb{±1.41}   & 56.17\sgb{±4.39} & 41.60\sgb{±4.55}  & 87.34\sgb{±4.60}   & 65.38\sgb{±2.36} & 50.78\sgb{±3.03}  & 88.07\sgb{±1.37}   \\
DRUM      & 86.83\sgb{±2.17} & 80.70\sgb{±3.63}  & 98.06\sgb{±1.31}   & 47.02\sgb{±5.08} & 30.11\sgb{±5.35}  & 83.76\sgb{±4.15}   & 68.88\sgb{±1.40} & 53.37\sgb{±1.22}  & 92.15\sgb{±1.20}   \\
RNNLogic  & 85.74\sgb{±2.56} & 76.40\sgb{±2.88}  & 96.18\sgb{±1.20}   & 61.78\sgb{±6.04} & 45.70\sgb{±6.83}  & 91.26\sgb{±5.24}   & 70.20\sgb{±4.92} & 58.40\sgb{±5.50}  & 91.04\sgb{±2.51}   \\		
RLogic    & 86.08\sgb{±2.71} & 75.81\sgb{±3.26}  & 97.27\sgb{±1.88}   & 59.70\sgb{±6.51} & 45.60\sgb{±6.87}  & 89.08\sgb{±5.12}   & 67.74\sgb{±7.17} & 54.32\sgb{±8.33}  & 92.55\sgb{±3.17}   \\
NCRL      & 87.18\sgb{±1.55} & 78.07\sgb{±2.89}  & 94.34\sgb{±3.25}   & 64.32\sgb{±7.39} & 51.39\sgb{±7.94}  & 91.56\sgb{±5.89}   & 74.73\sgb{±3.98} & 59.84\sgb{±5.44}  & 95.15\sgb{±3.78}   \\
\model       & \underline{88.88\sgb{±0.57}} & \underline{81.14\sgb{±1.33}}  & \underline{98.27\sgb{±0.49}}   & \underline{66.01\sgb{±5.69}} & \underline{52.12\sgb{±6.09}}  & \underline{92.14\sgb{±4.62}}   & \underline{75.90\sgb{±3.88}} & \underline{62.00\sgb{±5.06}}  & \underline{95.52\sgb{±3.91}}  \\
     \midrule[1.5pt]
     \midrule[1.5pt]

\multirow{2}{*}{Methods} & \multicolumn{3}{c}{WN-18RR}              & \multicolumn{3}{c}{FB15K-237}             & \multicolumn{3}{c}{YAGO3-10}                \\
& MRR (\%)   & HIT@1 (\%) & HIT@10 (\%) & MRR (\%)   & HIT@1 (\%) & HIT@10 (\%) & MRR (\%)   & HIT@1 (\%) & HIT@10 (\%) \\
\hline
TransE    & 19.05\sgb{±2.08} & 0.91\sgb{±0.82}   & 45.54\sgb{±6.13}   & 37.40\sgb{±1.63} & 24.65\sgb{±1.09}  & 61.84\sgb{±2.70}   & 34.42\sgb{±4.13} & 26.78\sgb{±5.14}  & 56.01\sgb{±4.53}   \\
DistMult  & 37.67\sgb{±2.44} & 34.06\sgb{±3.26}  & 45.38\sgb{±5.31}   & 35.53\sgb{±1.15} & 23.93\sgb{±1.24}  & 58.46\sgb{±1.59}   & 27.56\sgb{±3.16} & 19.54\sgb{±2.56}  & 42.93\sgb{±4.27}   \\
ComplEx   & 39.99\sgb{±3.67} & 36.35\sgb{±3.06}  & 47.15\sgb{±6.89}   & 37.49\sgb{±1.81} & 26.26\sgb{±1.52}  & 59.67\sgb{±2.64}   & 36.34\sgb{±2.45} & 27.23\sgb{±2.24}  & 53.86\sgb{±3.44}   \\
RotatE    & 41.17\sgb{±3.99} & 36.92\sgb{±3.23}  & 49.64\sgb{±7.00}   & 40.40\sgb{±1.99} & 28.24\sgb{±1.70}  & \textbf{63.48\sgb{±2.70}}   & \textbf{40.41\sgb{±3.87}} & \textbf{31.76\sgb{±4.34}}  & 56.70\sgb{±3.70}   \\
\hdashline

CompGCN  & 45.32\sgb{±3.45} & 39.25\sgb{±3.35} & 54.63\sgb{±4.53} &  33.74\sgb{±2.03} &22.98\sgb{±1.73} &52.55\sgb{±2.21} &28.43\sgb{±2.12} &21.36\sgb{±3.67} &47.52\sgb{±3.38}  \\
NBFNet   & \textbf{53.87\sgb{±4.75}} & \textbf{49.50\sgb{±3.04}}  & 62.93\sgb{±4.94} & 39.22\sgb{±2.33} & 28.79\sgb{±2.43}  & 57.39\sgb{±3.13}   & 36.83\sgb{±2.31} & 27.91\sgb{±4.16} & 55.42\sgb{±3.24}     \\
A*Net   & 51.18\sgb{±4.15} & 42.82\sgb{±4.45} & 64.37\sgb{±5.05} & 39.88\sgb{±2.45} & 30.89\sgb{±2.76} & 58.13\sgb{±2.03}  & 36.16\sgb{±4.08} & 27.36\sgb{±4.31} & 55.93\sgb{±3.33} \\

\hdashline
Neural-LP & 37.67\sgb{±5.34} & 28.46\sgb{±5.13}  & 56.15\sgb{±5.46}   & 31.45\sgb{±1.53} & 22.71\sgb{±1.35}  & 48.56\sgb{±1.93}   & -          & -           & -            \\
DRUM      & 37.68\sgb{±5.46} & 28.41\sgb{±5.28}  & 56.30\sgb{±5.38}   & 31.89\sgb{±1.42} & 23.17\sgb{±1.28}  & 48.75\sgb{±1.87}   & -          & -           & -            \\
RNNLogic  & 45.28\sgb{±5.35} & 39.00\sgb{±5.32}  & 55.32\sgb{±5.27}   & 28.59\sgb{±1.65} & 21.20\sgb{±1.74}  & 45.62\sgb{±1.37}   & -          & -           & -            \\
RLogic    & 46.28\sgb{±6.42} & 37.64\sgb{±6.08}  & 65.22\sgb{±4.84}   & 29.00\sgb{±2.23} & 19.78\sgb{±1.93}  & 44.02\sgb{±3.21}   & 35.88\sgb{±4.91} & 27.22\sgb{±3.24}  & 52.25\sgb{±4.64}   \\
NCRL      & 49.02\sgb{±6.35} & 41.44\sgb{±7.68}  & 64.79\sgb{±4.41}   & 38.76\sgb{±3.81} & 29.24\sgb{±3.98}  & 58.46\sgb{±6.68}   & 38.33\sgb{±4.93} & 29.87\sgb{±5.49}  & 56.14\sgb{±4.68}   \\
\model       & \underline{51.01\sgb{±5.97}} & \underline{43.44\sgb{±5.84}}  & \underline{\textbf{65.67\sgb{±4.22}}}   & \underline{\textbf{41.16\sgb{±2.42}}} & \underline{\textbf{31.22\sgb{±2.17}}}  & \underline{61.29\sgb{±3.01}}   &   \underline{40.26\sgb{±4.29}}   &   \underline{31.19\sgb{±4.51}}  &  \underline{\textbf{57.12\sgb{±3.01}}}  \\         
     \bottomrule[2pt]
\end{tabular}}
\label{table_subpop}
\end{table*}

%% file: section/table_no_shift.tex
\begin{table}[htb]
\centering
\renewcommand\arraystretch{1.2}
\caption{Knowledge graph completion results for rule learning methods on FB15K-237 and YAGO3-10 without synthetic query shifts.}
\resizebox{\linewidth}{!}{%
\begin{tabular}{ccccccc}
 \toprule[2pt]
\multirow{2}{*}{Methods} & \multicolumn{3}{c}{FB15K-237}              & \multicolumn{3}{c}{YAGO3-10}             \\
& MRR  & HIT@1& HIT@10 & MRR   & HIT@1 & HIT@10 \\
\hline    
Neural-LP & 31.56 &	22.84 & 49.32 & - & - & - \\
DRUM      & 31.87 & 23.29 & 48.92 & - & - & - \\
RNNLogic  & 28.67 & 21.35 & 45.73 & - & - & - \\	
RLogic  & 29.55  & 19.87  & 44.26  & 36.41  & 25.87  & 50.38 \\
NCRL      & 39.28  & 29.54  & 60.06  & 38.26  & 29.67  & 56.16\\
\model       & \textbf{41.45}  & \textbf{31.20}  & \textbf{61.59}  & \textbf{39.89}  & \textbf{31.06}  & \textbf{57.35}\\
 \bottomrule[2pt]
\end{tabular}}
\label{table_no_shift}
\end{table}

%% file: section/table_inductive.tex
\begin{table*}
\centering
\renewcommand\arraystretch{1.2}
\caption{Inductive link prediction results across three datasets, averaged over five subgroups. Bold and underlined values denote the best results for each dataset among all methods and rule-based methods, respectively.}
\resizebox{\linewidth}{!}{%
\begin{tabular}{cccccccccc}
 \toprule[2pt]
\multirow{2}{*}{Methods} & \multicolumn{3}{c}{WN-18RR}              & \multicolumn{3}{c}{FB15K-237}             & \multicolumn{3}{c}{YAGO3-10}                \\
 & MRR (\%)   & HIT@1 (\%) & HIT@10 (\%) & MRR (\%)   & HIT@1 (\%) & HIT@10 (\%) & MRR (\%)   & HIT@1 (\%) & HIT@10 (\%) \\
\hline
KGEs 
 & - & -   & -   & - & -  & -   & - & -  & -   \\
\hdashline

NBFNet   & \textbf{52.96\sgb{±4.43}} & \textbf{48.63\sgb{±3.25}}  & 61.39\sgb{±4.11}   
& 37.91\sgb{±1.89} & 26.36\sgb{±2.98}  & 56.47\sgb{±3.77}  & 36.15\sgb{±2.13} & 26.28\sgb{±2.98}  & 56.47\sgb{±3.77}   \\
A*Net   & \textbf{50.88\sgb{±3.88}} & \textbf{42.33\sgb{±3.06}}  & 63.79\sgb{±4.00}   
& 37.41\sgb{±1.93} & 26.02\sgb{±2.74}  & 57.05\sgb{±2.99}  & 34.86\sgb{±2.78} & 25.75\sgb{±2.77}  & 55.29\sgb{±3.12}  \\ 

\hdashline

    Neural-LP & 24.14\sgb{±2.34} & 19.86\sgb{±2.57} & 35.56\sgb{±3.82} & 14.38\sgb{±1.71} & 9.10\sgb{±1.83}  & 29.34\sgb{±2.13} & -          & -          & -          \\
 DRUM      & 25.58\sgb{±2.36} & 21.06\sgb{±2.42} & 38.16\sgb{±3.85} & 17.25\sgb{±1.62} & 11.70\sgb{±1.72} & 30.06\sgb{±2.28} & -          & -          & -          \\
 RNNLogic  & 31.45\sgb{±3.88} & 26.00\sgb{±5.40} & 45.36\sgb{±4.73} & 25.40\sgb{±1.76} & 19.40\sgb{±2.04} & 40.74\sgb{±1.83} & -          & -          & -          \\
 RLogic    & 41.06\sgb{±4.90} & 37.04\sgb{±5.54} & 51.24\sgb{±4.87} & 26.15\sgb{±1.93} & 19.78\sgb{±2.37} & 43.80\sgb{±3.26} & 31.29\sgb{±2.91} & 23.89\sgb{±3.32} & 49.74\sgb{±2.84} \\
 NCRL      & 45.46\sgb{±5.57} & 39.64\sgb{±6.24} & 61.20\sgb{±4.96} & 35.53\sgb{±3.40} & 26.43\sgb{±3.38} & 54.30\sgb{±4.83} & 33.61\sgb{±3.36} & 27.44\sgb{±4.40} & 50.86\sgb{±3.12} \\
 \model       & \underline{48.53\sgb{±4.25}} & \underline{42.44\sgb{±4.94}} & \underline{\textbf{63.68\sgb{±4.64}}} & \underline{\textbf{38.86\sgb{±2.57}}} & \underline{\textbf{29.92\sgb{±3.01}}} & \underline{\textbf{60.08\sgb{±3.07}}} & \underline{\textbf{38.18\sgb{±2.57}}} & \underline{\textbf{30.09\sgb{±2.83}}} & \underline{\textbf{56.28\sgb{±2.54}}}\\
     \bottomrule[2pt]
\end{tabular}}
\label{table_ind}
\end{table*}

%% file: section/table_rules.tex
\begin{table*}
\renewcommand\arraystretch{1}
\centering
\caption{Top rules learned by \model~on YAGO3-10.}
\resizebox{0.75\linewidth}{!}{
\begin{tabular}{c}
 \toprule[2pt]
 isLocatedIn$(x,y) \leftarrow $hasNeighbor$(x,z) \wedge $isLocatedIn$(z,y)$
  \\
  isLocatedIn$(x,y) \leftarrow $hasNeighbor$(x,z_1) \wedge $ dealsWith$(z_1,z_2) \wedge $ isLocatedIn$(z_2,y)$
\\ \hline
 isCitizenOf$(x,y) \leftarrow $isMarriedTo$(x,z) \wedge $isCitizenOf$(z,y)$
\\
  isCitizenOf$(x,y) \leftarrow $hasAcademicAdvisor$(x,z_1) \wedge $ worksAt$(z_1,z_2) \wedge $ isLocatedIn$(z_2,y)$
\\ \hline
 livesIn$(x,y) \leftarrow $worksAt$(x,z) \wedge $isLocatedIn$(z,y)$
\\
livesIn$(x,y) \leftarrow $hasChild$(x,z_1) \wedge $ isMarriedTo$(z_1,z_2) \wedge $ livesIn$(z_2,y)$
\\ \hline
 graduatedFrom$(x,y) \leftarrow $hasAcademicAdvisor$(x,z) \wedge $worksAt$(z,y)$
\\ 
graduatedFrom$(x,y) \leftarrow $hasAcademicAdvisor$(x,z_1) \wedge $ influences$(z_1,z_2) \wedge $ graduatedFrom$(z_2,y)$
\\
\bottomrule[2pt]
\end{tabular}
}
\label{table_rule}
\end{table*}

%% file: section/conclusion.tex
\section{Further Studies}
\subsection{Convergence Property}

Figure~\ref{figure_loss} presents the training loss trajectories across all six datasets used in KGC experiments. The fluctuations during intermediate training stage is because our model learn rules of different length sequentially: it first learns 2-length rules before advancing to 3-length rules.
We observe two findings: First, larger batch sizes (corresponding to training on larger KGs) yield more stable training trajectories with reduced fluctuation magnitudes. Second, all learning curves demonstrate convergence prior to reaching the prescribed iteration limit, with no empirical evidence of improved performance from additional training. This convergence behavior suggests the feasibility of implementing an early stopping criterion without compromising model performance.

\subsection{Mined Rules}

Table~\ref{table_rule} presents the logical rules learned by \model~on the YAGO3-10 dataset. These rules are characterized by their interpretability and rigorous structure, offering a faithful explanation for each head relation.

We also carefully compare the rules learned by NCRL and \model, revealing two advantages of our approach. 
First, regarding rule generality, table~\ref{table_rule_comp} demonstrates that NCRL predominantly relies on education-related information for both districts and individuals. In contrast, our model captures more fundamental and invariant properties: neighborhood relationships for districts, and professional as well as familial connections for individuals. 
Second, in terms of semantic validity, we observe that NCRL occasionally learns rules with questionable applicability, such as associating the \textit{hasAcademicAdvisor} relation with districts - an implausible linkage that generate false predictions. By comparison, our model consistently generates semantically appropriate rules that properly reflect domain knowledge and entity-relation compatibility.

\input{section/table_rules_compare}

\subsection{Decorrelation Visualization}

\begin{figure}[htb]
\centering
\includegraphics[width=\linewidth]{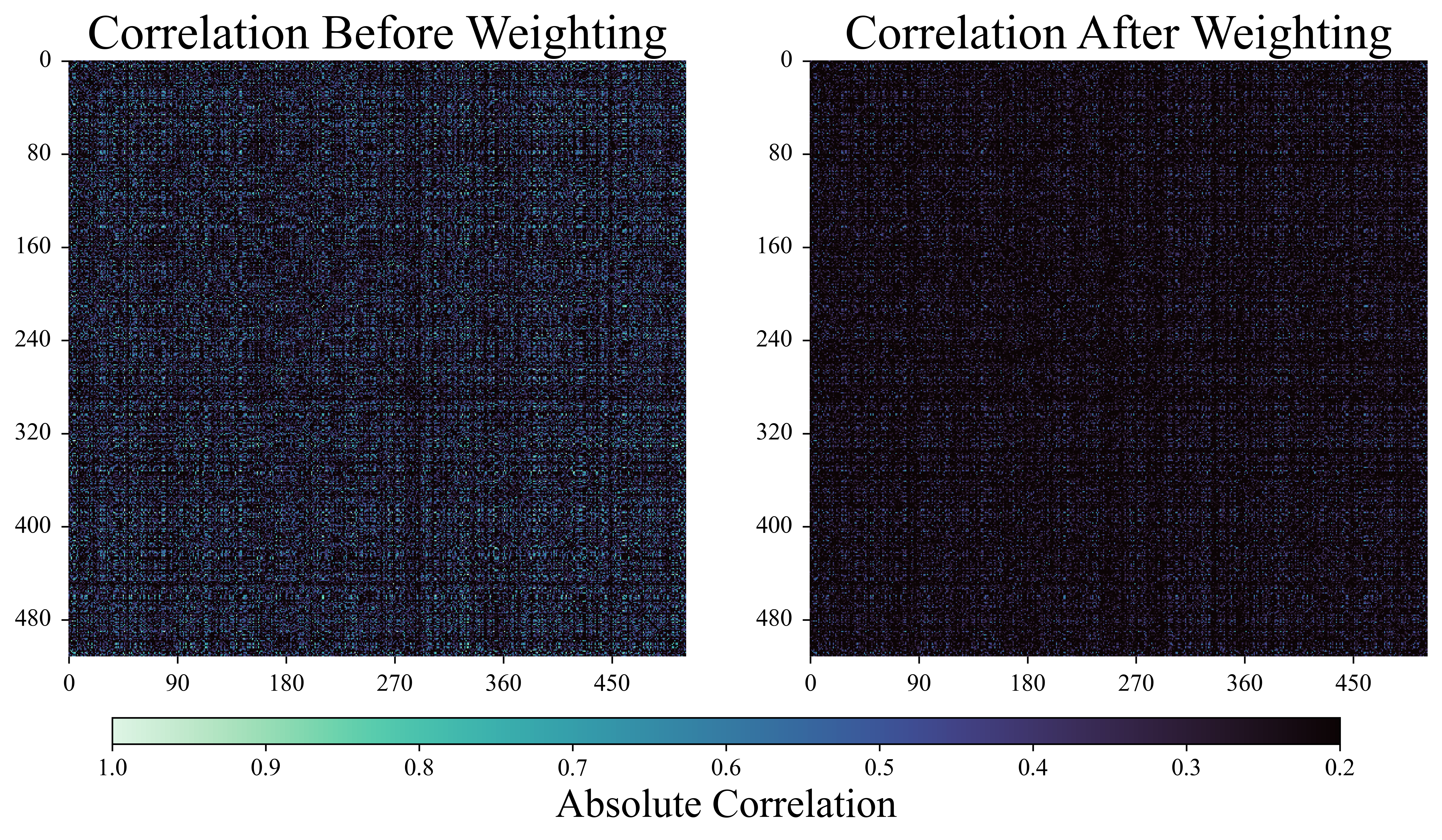}
\caption{Comparison of feature correlation matrices before (left) and after (right) reweighting. Diagonal elements are set to zero. The reweighted version exhibits reduced inter-feature correlations.}
\label{figure_corr}
\end{figure}
\input{section/table_ablation}

We compare the feature correlation matrices before and after reweighting. To isolate inter-feature correlations, we explicitly zero out the diagonal elements, as self-correlations are uninformative for measuring independence. As shown in Figure~\ref{figure_corr}, the left image (before decorrelation) exhibits distinct banded and grid-like correlation patterns, whereas the right image (after decorrelation) appears nearly blank, with significantly reduced values. This visual result demonstrates that our decorrelation module successfully removes spurious feature correlations through sample reweighting, thereby supporting the effective learning of plausibility scores.

\subsection{Ablation Stuidies}
Since our model integrates feature decorrelation and rule learning within a unified framework, this section presents ablation studies to assess the utility and sensitivity of its key components and design choices. We examine three critical aspects: (1) feature decorrelation (ablation variant termed NoDecor), (2) higher-order moment dependencies elimination (variant that only reduces first-order moment dependencies termed FirstMoment), and (3) the backtracking mechanism for efficient rule instance generation (variant without this component termed NoBack).  These configurations are evaluated on FB15K-237 and YAGO3-10 datasets using the same KGC settings as in previous section.

The results in Table~\ref{table_ablation} reveal three key findings. First, removing feature decorrelation causes an average 5.18\% MRR performance drop, validating the importance of this regularization component. Second, while the FirstMoment variant outperforms complete regularizer removal, it still underperforms compared to the full model. Third, disabling the backtracking mechanism results in an average 2.99\% MRR reduction, indicating that generating diverse rule instances also helps address distribution shift.

\subsection{Efficiency Analysis}

To evaluate the scalability of \model, we compare its training time against other rule-learning methods on the FB15k-237 dataset. While not the largest KG in our study, FB15k-237 is the most relationally dense and remains computationally tractable for all baseline methods. Empirical results show favorable efficiency: NeuralLP requires 23,715 seconds, DRUM takes 22,568 seconds, and RNNLogic fails to converge within a day. In contrast, NCRL completes training in 445 seconds, while \model~achieves better performance at 884 seconds. The marginal overhead of \model~stems from its decorrelation modules, which mitigate covariate shifts at a negligible computational cost.

Theoretically, the time complexity of \model~is $\mathcal{O}\big((\frac{T_d}{d}O_d^2 + 2)N^+d^2\big)<\mathcal{O}\big((O_d^2 + 2)N^+d^2\big)$, where $T_d$ denotes the number of weight learning epochs, $O_d$ represents the moment order parameter in Equation~\eqref{eqa4} (with $O_d = 2$ in our experiments), $N^+$ corresponds to the number of sampled paths, and $d$ is the representation dimensionality. In comparison, NeuralLP scales as $\mathcal{O}\big(|R|^l|E|^{3(l-1)}\big)$ with $|R|$, $|E|$, and $l$ representing the number of relations, entities, and rule length respectively, while NCRL achieves $\mathcal{O}\big(2N^+d^2\big)$. The above analysis demonstrates that \model~maintains superior computational efficiency compared to all rule-learning methods except NCRL, while maintaining robust generalization capabilities.



\section{Conclusion}

In this work, we investigate the challenge of learning logical rules for KG reasoning under agnostic distribution shifts. We propose a stable rule learning framework comprising two key components: (1) a feature decorrelation regularizer and (2) a rule-learning network. Our novel framework employs a re-weighting mechanism designed to decorrelate rule body embeddings, effectively addressing the covariate shift problem induced by query shifts. Comprehensive experiments across different reasoning settings demonstrate the efficacy of our method.


Our ongoing research will pursue three key directions: First, we aim to further reduce both linear and non-linear statistical dependencies in embeddings through Random Fourier Feature analysis. Second, we intend to systematically investigate additional sources of distribution shifts and develop corresponding mitigation strategies. Finally, we will explore the development of stable KGE models for handling agnostic distribution shifts, potentially trading some interpretability for enhanced performance.

%% file: section/table_rules_compare.tex
\begin{table*}
\renewcommand\arraystretch{1.1}
\caption{Examining Rules learned by our framework and NCRL}
\resizebox{\linewidth}{!}{
\begin{tabular}{ccc}
 \toprule[2pt]
 Relation & \model & NCRL \\ \hline
 \multirow{2}{*}{isLocatedIn} & hasNeighbor$(x,z) \wedge $isLocatedIn$(z,y)$ &
 isLocatedIn$(x,z) \wedge $isLocatedIn$(z,y)$
  \\
  &
  hasNeighbor$(x,z_1) \wedge $ dealsWith$(z_1,z_2) \wedge $ isLocatedIn$(z_2,y)$ &
  hasAcademicAdvisor$(x,z_1) \wedge $ isLocatedIn$(z_1,z_2) \wedge $ isLocatedIn$(z_2,y)$
  \\ \hline

 \multirow{2}{*}{livesIn} & worksAt$(x,z) \wedge $isLocatedIn$(z,y)$ &
 graduatedFrom$(x,z) \wedge $isLocatedIn$(z,y)$
  \\
  & 
  hasChild$(x,z_1) \wedge $ isMarriedTo$(z_1,z_2) \wedge $ livesIn$(z_2,y)$ &
  hasAcademicAdvisor$(x,z_1) \wedge $ hasAcademicAdvisor$(z_1,z_2) \wedge $ livesIn$(z_2,y)$
\\ 
\bottomrule[2pt]
\end{tabular}
}
\label{table_rule_comp}
\end{table*}

%% file: section/table_ablation.tex
\begin{table}[tb]
\centering
\renewcommand\arraystretch{1.2}
\caption{Mean Performance in Ablation Studies on FB15K-237 and YAGO3-10}
\resizebox{\linewidth}{!}{%
\begin{tabular}{ccccccc}
 \toprule[2pt]
\multirow{2}{*}{Methods} & \multicolumn{3}{c}{FB15K-237}              & \multicolumn{3}{c}{YAGO3-10}             \\
& MRR  & HIT@1& HIT@10 & MRR   & HIT@1 & HIT@10 \\
\hline    

\model       & 41.16  & 31.22  & 61.29  & 40.26  & 31.19  & 57.12\\
$\rhd$NoDecor & 38.91 &	29.43 & 59.13 & 38.29 & 29.73 & 56.44 \\
$\rhd$FirstMoment & 40.37 & 30.14 & 59.56 & 39.40 & 30.64 & 56.92 \\
$\rhd$NoBack  & 39.86 &	29.77 & 59.24 & 39.12 & 30.31 & 56.83 \\
 \bottomrule[2pt]
\end{tabular}}
\label{table_ablation}
\end{table}

%% file: section/biography.tex
 


\begin{IEEEbiography}[{\includegraphics[width=1in,height=1.25in,clip,keepaspectratio]{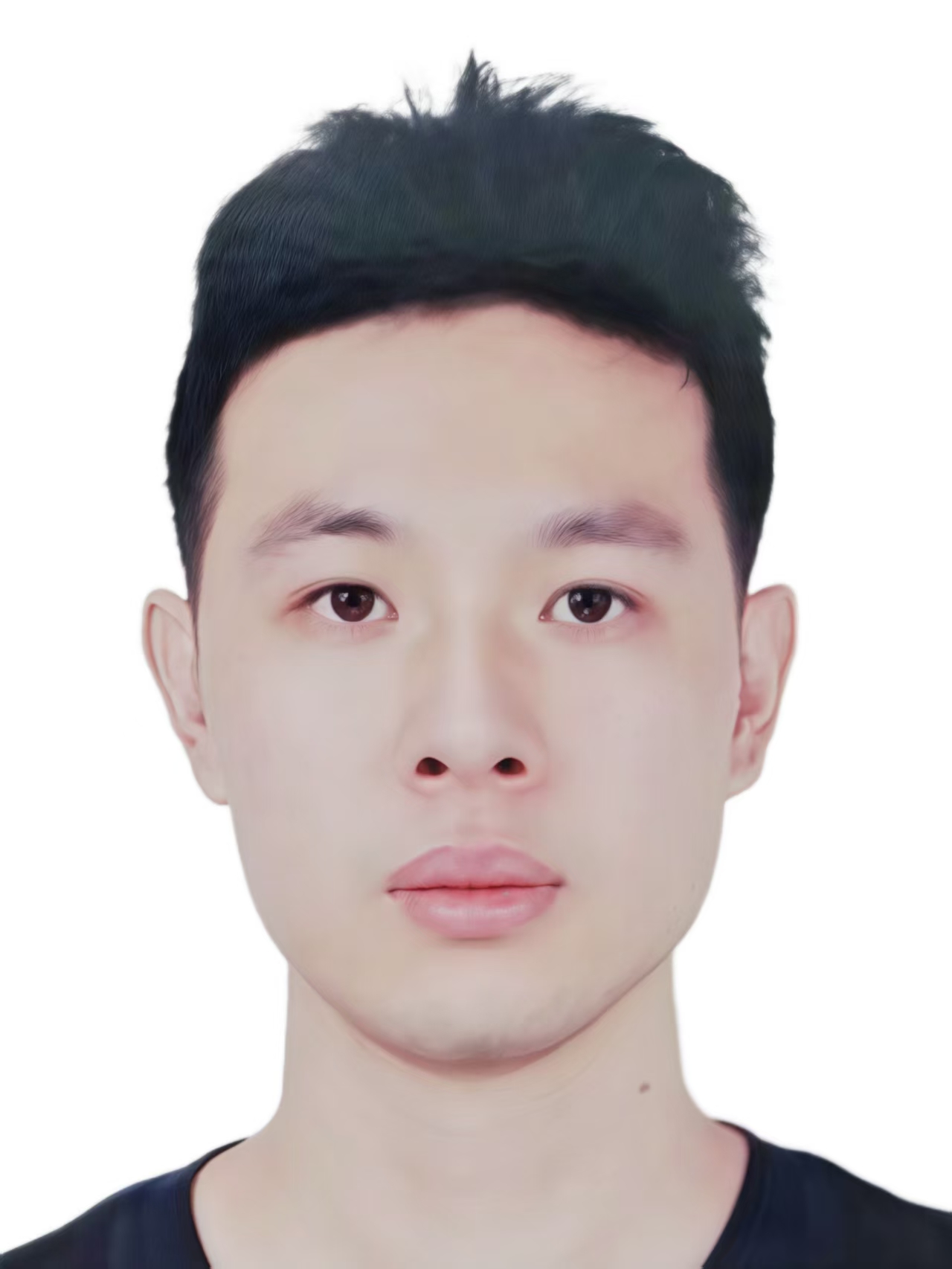}}]
{Shixuan Liu}
received his B.S. and Ph.D. degrees from the National University of Defense Technology, Changsha, China, in 2019 and 2024, respectively. He is also a visiting scholar in the Department of Computer Science and Technology at Tsinghua University, where he has spent two years. He has published over 10 papers in prestigious journals and conferences, including T-PAMI, T-KDE, T-CYB, CIKM and ICDM, focusing on knowledge reasoning and data mining.
\end{IEEEbiography}

\begin{IEEEbiography}[{\includegraphics[width=1in,height=1.25in,clip,keepaspectratio]{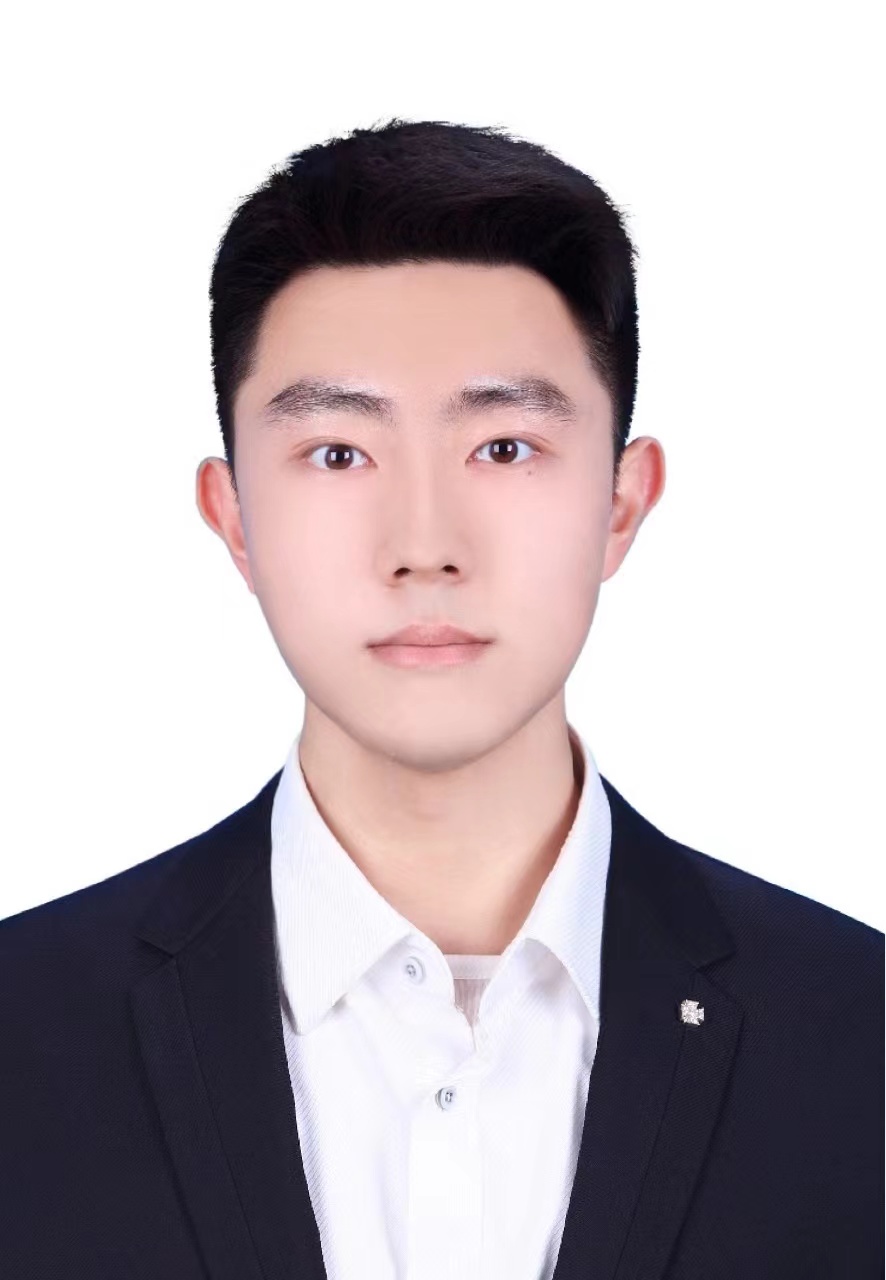}}]{Yue He}
is a Assistant Professor at School of Information, Renmin University of China. He received his Ph.D. in the Department of Computer Science and Technology from Tsinghua University in 2023. His research interests include out-of-distribution generalization, causal structure learning, and graph computing. He has published more than 20 papers in prestigious conferences and journals in machine learning, data mining, and computer vision. He serves as a PC member in many academic conferences, including ICML2024, Neurips2024, UAI2024 and etc.
\end{IEEEbiography}

\begin{IEEEbiography}[{\includegraphics[width=1in,height=1.25in,clip,keepaspectratio]{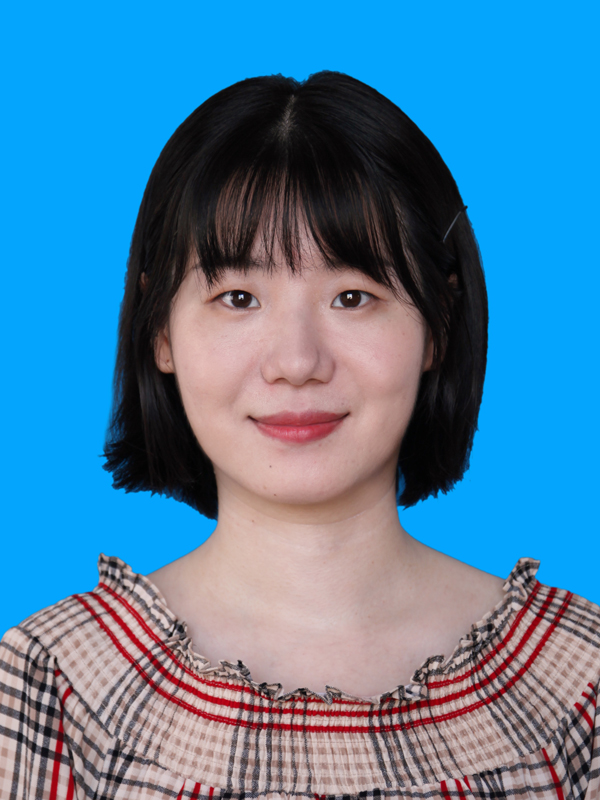}}]{Yunfei Wang}
received the B.S. degree in civil
engineering from the Hunan University, Changsha, China, in 2020. She is now pursuing the Ph.D degree at the National University of Defense Technology, Changsha, China. Her research interests include auto penetration test, reinforcement learning and cyber-security.
\end{IEEEbiography}

\begin{IEEEbiography}[{\includegraphics[width=1in,height=1.25in,clip,keepaspectratio]{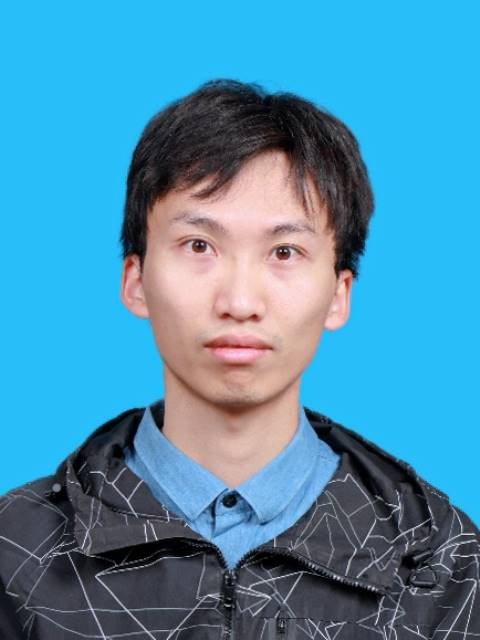}}]{Hao Zou} received the B.E. degree and Ph.D. degree from the Department of Computer Science and technology, Tsinghua University respectively in 2018 and 2023. His main research interests including causal inference and counterfactual learning.
\end{IEEEbiography}

\begin{IEEEbiography}[{\includegraphics[width=1in,height=1.25in,clip,keepaspectratio]{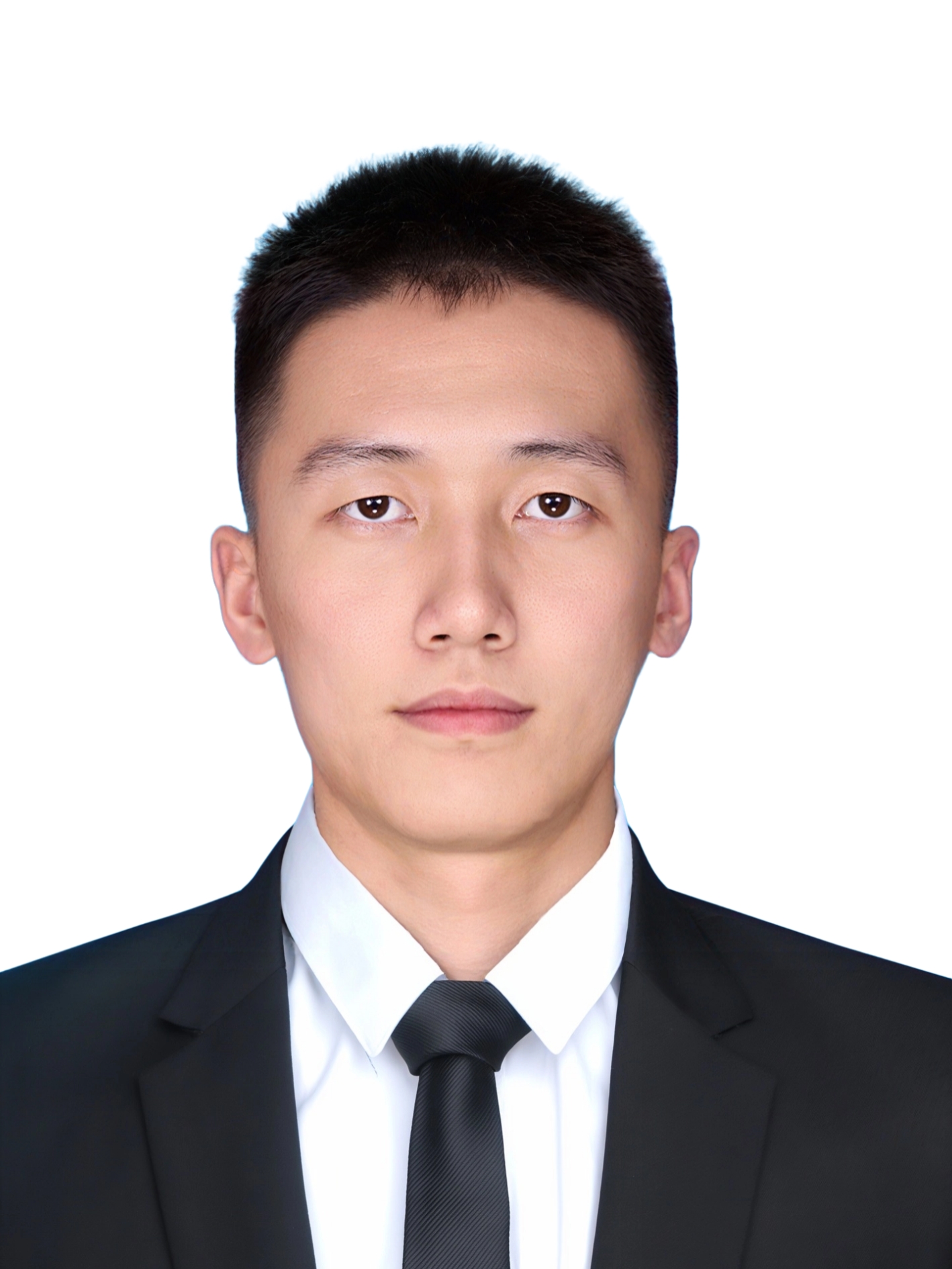}}]{Haoxiang Cheng} received the B.S. degree in systems engineering from the National University of Defense Technology, Changsha, China, in 2024, where he is currently pursuing the master's degree. His research interests include large language models, and knowledge reasoning.
\end{IEEEbiography}

\begin{IEEEbiography}[{\includegraphics[width=1in,height=1.25in,clip,keepaspectratio]{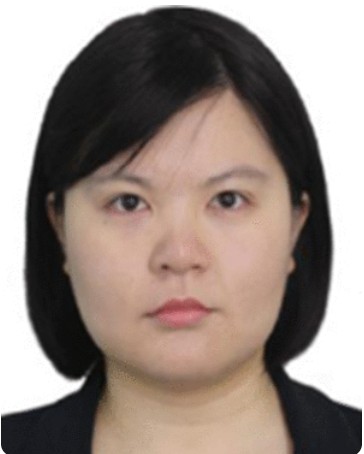}}]{Wenjing Yang} received the PhD degree in multi-scale modelling from Manchester University, Manchester, U.K., in 2014. She is currently a research fellow with the College
of Computer Science and Technology, National University of Defense Technology. Her research interests include machine learning, robotics software, high-performance computing and causal inference.

\end{IEEEbiography}

\begin{IEEEbiography}[{\includegraphics[width=1in,height=1.25in,clip,keepaspectratio]{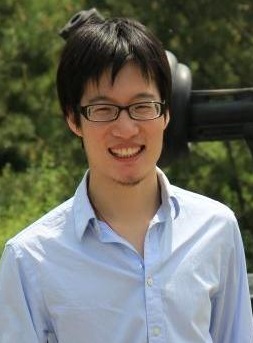}}]{Peng Cui}
(Member, IEEE) is an Associate Professor with tenure in Tsinghua University. He got his PhD degree from Tsinghua University in 2010. His research interests include causally-regularized machine learning, network representation learning, and social dynamics modeling. He has published more than 100 papers in prestigious conferences and journals in data mining and multimedia. Now his research is sponsored by National Science Foundation of China, Samsung, Tencent, etc. He also serves as guest editor, co-chair, PC member, and reviewer of several high-level international conferences, workshops, and journals.
\end{IEEEbiography}

\begin{IEEEbiography}[{\includegraphics[width=1in,height=1.25in,clip,keepaspectratio]{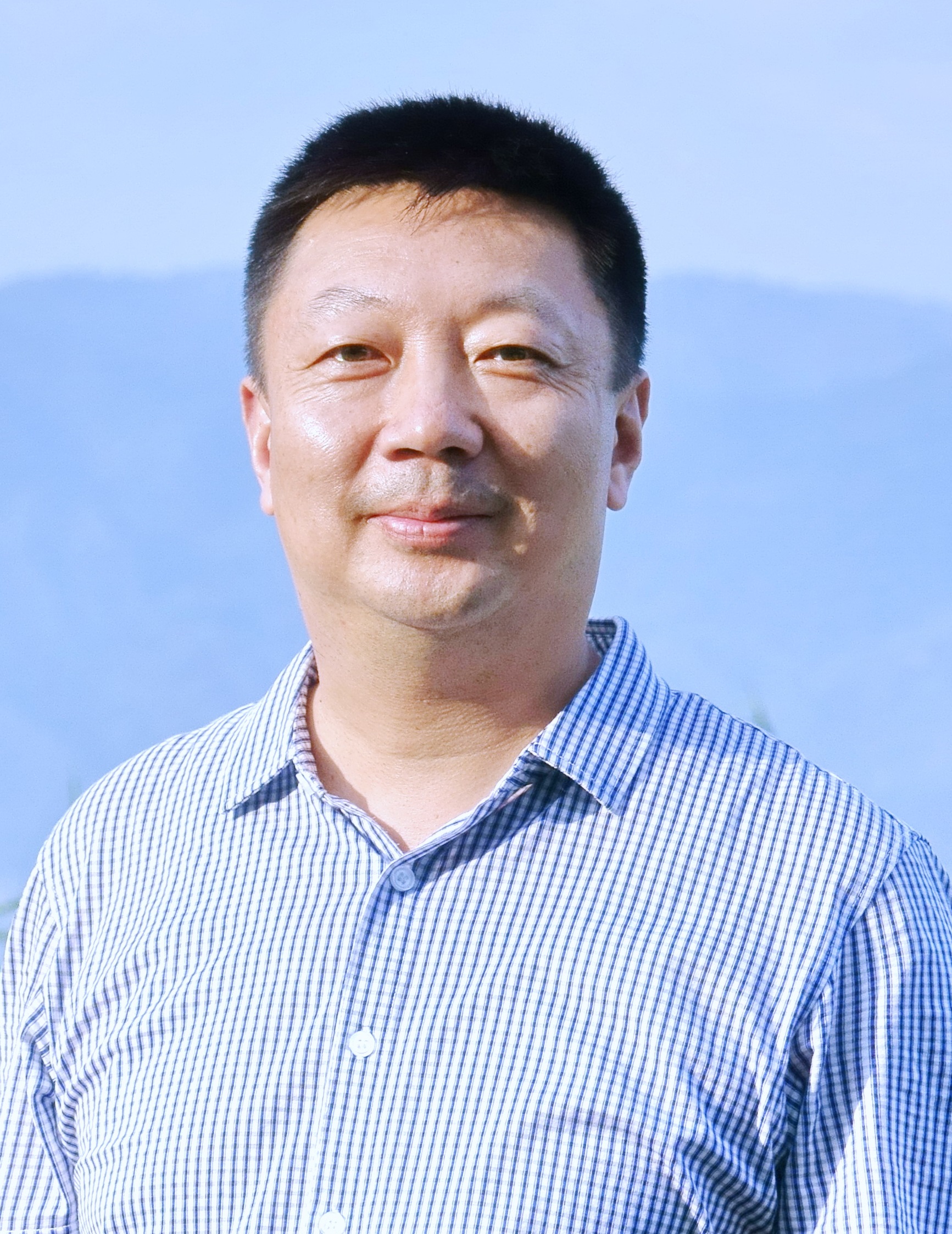}}]{Zhong Liu}
received the B.S. degree in Physics from Central China Normal University, Wuhan, Hubei, China, in 1990, the M.S. degree in computer software and the Ph.D. degree in management science and engineering both from National University of Defense Technology, Changsha, China, in 1997 and 2000. He is a professor in the College of Systems Engineering, National University of Defense Technology, Changsha, China. His research interests include intelligent information systems, and intelligent decision making.
\end{IEEEbiography}

\vfill

%% file: section/appendix.tex
\clearpage
\appendices
\input{section/table_hyperparameters}

\section{Dataset Description}
\label{sec:dataset}
\begin{itemize}
\item \textbf{Family} captures family relations among family members~\cite{hinton1986learning}.
\item \textbf{Kinship} contains kinship relations among members of the Alyawarra tribe from
Central Australia~\cite{kok2007statistical}
\item \textbf{UMLS} originates from the field of bio-medicine. It contains entities that are biomedical concepts, and relations such as treatments and diagnoses~\cite{kok2007statistical}
\item \textbf{WN18RR} is designed to function as a user-friendly dictionary and thesaurus, as well as facilitate automatic text analysis. Its entities correspond to word senses,
and relations define lexical connections between them~\cite{dettmers2018convolutional}.
\item \textbf{FB15K-237} represents an online collection of structured data extracted from numerous sources, including user-submitted Wiki~\cite{toutanova2015observed}.

\item \textbf{YAGO3-10} is a subset of YAGO, a large semantic knowledge base, derived from a variety of sources, such as Wikipedia, WordNet, WikiData, GeoNames~\cite{suchanek2007yago}.
\item \textbf{DBLP} is a citation network dataset is extracted from DBLP, ACM, Microsoft Academic Graph, and other sources, seeking to understand the relations between scientific papers, authors and publishing venues~\cite{tang2008arnetminer}
\end{itemize}    

\section{Baseline Description}
\label{sec:baseline_appendix}
\begin{itemize}
\item \textbf{TransE}~\cite{bordes2013transe}: A foundational knowledge graph embedding method that represents relations as translations between entities in a low-dimensional space.
\item \textbf{DistMult}~\cite{yang2014embedding}: A bilinear model that captures pairwise interactions between entities using diagonal relation matrices.
\item \textbf{ComplEx}~\cite{trouillon2016complex}: Extends DistMult to complex-valued embeddings to model both symmetric and antisymmetric relations.
\item \textbf{RotatE}~\cite{sun2019rotate}: Represents relations as rotations in complex space to model various relation patterns including composition.
\item \textbf{CompGCN}~\cite{vashishthRGCN}: A GNN that jointly embeds nodes and relations using composition operations for neighborhood aggregation.
\item \textbf{NBFNet}~\cite{zhu2021neural}: Generalizes path-based methods by learning to dynamically combine paths for relation prediction.
\item \textbf{A* Net}~\cite{zhu2023net}: Extends NBFNet with A*-like heuristic search to guide the reasoning process.
\item \textbf{Neural-LP}~\cite{yang2017differentiable}: A differentiable approach to learning logical rules using tensor operations.
\item \textbf{DRUM}~\cite{sadeghian2019drum}: Extends Neural-LP with RNNs to handle long rule chains and prune incorrect rules.
\item \textbf{RNNLogic}~\cite{qu2020rnnlogic}: Uses an RNN-based generator to produce interpretable logical rules separately apart from reasoning.
\item \textbf{RLogic}~\cite{cheng2022rlogic}: Introduces recursive logical rule learning through predicate representation learning.
\item \textbf{NCRL}~\cite{cheng2022neural}: A neural compositional rule learning method using attention mechanisms for recursive reasoning.
\end{itemize}

\section{Forward Chaining}
\label{sec:chain}
By using rules $\mathcal{R}$ and rule score, the score for each candidate answer could be efficiently derived from logical rules with forward chaining after path counting~\cite{salvat1996sound}. Thus, the final predicted answer is obtained by selecting the candidate answer with the highest score. With matrix operations, this could be efficiently achieved by,
\begin{equation}
a=\mathop{argmax} \sum_{(\mathbf{r_b}, r_h)\in \mathcal{R}} s(\mathbf{r_b}, r_h) \prod_{r\in \mathbf{r_b}} \mathbf{M}_r \mathbf{V}_{e_h}
\end{equation}
where $\mathbf{M}_r \in \{0, 1\}^{|E|\times|E|}$  
describes the entity pairs connected by relation $r$ and $\mathbf{V}_{e_h}$ is the one-hot encoding for the head entity.

\section{Deductive Nature}
\label{sec:deductive}


Recent studies have emphasized the critical role of deductive reasoning in constructing effective rule body representations. This deductive property fundamentally means that the semantics of a complete logical expression are determined by both the meanings of its constituent atoms and their compositional structure. Furthermore, the order of deduction has been shown to significantly influence the fusion of neighboring relations within a rule body~\cite{cheng2022rlogic, sun2022heterogeneous}.

The challenge lies in the combinatorial complexity of determining the optimal deduction order. For a rule body $\mathbf{r_b} = [r_{b_1}, \cdots, r_{b_l}]$, there exist $\prod_{j=2}^{l-1}\frac{l+j-1}{j}$ possible decomposition paths. Given the prohibitive computational cost of identifying the globally optimal order, a greedy selection strategy that sequentially choose the most promising relation pair at each step is preferable.

\section{Hyper-Parameters}
The hyper-Parameters for our method are listed in Table~\ref{table_hyperparamenters}

%% file: section/table_hyperparameters.tex
\begin{table*}[tb]
\centering
\caption{Hyper-parameters}\label{table_hyperparamenters}
\resizebox{\linewidth}{!}{%
\begin{tabular}{c|c|c|p{0.3\linewidth}}
\toprule[1.5pt]
\textbf{Hyper-parameter}& \textbf{Symbol}& \textbf{Value} & \multicolumn{1}{c}{\textbf{Description}}
\\ \midrule[1.2pt]
\multirow{4}{*}{Batch Size} &\multirow{4}{*}{$N$}                              & 500 (Family)       & \multirow{4}{*}{Batch Size for Rule Samples}                   \\
&&1000 (Kinship, UMLS, DBLP)&\\
&&5000 (WN18-RR, FB15K-237)&\\ 
&&8000 (YAGO3-10)&\\
\hline
Rule Length &$l$                              & 3            & Maximum length for Logical Rules       \\
\hline
\multirow{2}{*}{Embedding Size}   &\multirow{2}{*}{$d$}                              & 512 (Family, UMLS, DBLP)           & \multirow{2}{*}{Dimension of embeddings for relations}                         \\
&&1024 (Kinship, WN18-RR, FB15K-237, YAGO3-10)&
\\
\hline
\multirow{3}{*}{Rule Learning Epochs}    &\multirow{3}{*}{$T_r$}                       & 1000 (Family)         & \multirow{3}{*}{Overall Training Epochs}
\\
&&2000 (Kinship, UMLS, WN18-RR, DBLP)& 
\\
&&5000 (FB15K-237, YAGO3-10)& 
\\
\hline
\multirow{3}{*}{Rule Learning Rate}     &\multirow{3}{*}{$\alpha_r$}               &  0.00025 (Kinship, UMLS, YAGO3-10, DBLP) & \multirow{3}{*}{Learning rate to optimize Equation~\eqref{equ:overall}} \\
&& 0.0001 (Family, WN18-RR)&  \\
&& 0.0005 (FB15K-237)& \\
\hline
Weight Learning Rate &$\alpha_d$                              & 0.01            & Learning rate to optimize Equation~\eqref{eqa4}       \\
\hline
\multirow{3}{*}{Weight Learning Epochs}     &\multirow{3}{*}{$T_d$}               &  50 (WN-18RR) & \multirow{3}{*}{Re-sampling Weights Training Epochs} \\
&& 100 (UMLS, FB15K-237, YAGO3-10)&  \\
&& 200 (Family, Kinship)&  
\\
\hline
Weight Learning Order &$O_d$                              & 2            & The order in  Equation~\eqref{eqa4}       \\
\bottomrule[1.5pt]
\end{tabular}
}%
\end{table*}

%% file: main.bbl
\begin{thebibliography}{10}
\providecommand{\url}[1]{#1}
\csname url@samestyle\endcsname
\providecommand{\newblock}{\relax}
\providecommand{\bibinfo}[2]{#2}
\providecommand{\BIBentrySTDinterwordspacing}{\spaceskip=0pt\relax}
\providecommand{\BIBentryALTinterwordstretchfactor}{4}
\providecommand{\BIBentryALTinterwordspacing}{\spaceskip=\fontdimen2\font plus
\BIBentryALTinterwordstretchfactor\fontdimen3\font minus \fontdimen4\font\relax}
\providecommand{\BIBforeignlanguage}[2]{{%
\expandafter\ifx\csname l@#1\endcsname\relax
\typeout{** WARNING: IEEEtran.bst: No hyphenation pattern has been}%
\typeout{** loaded for the language `#1'. Using the pattern for}%
\typeout{** the default language instead.}%
\else
\language=\csname l@#1\endcsname
\fi
#2}}
\providecommand{\BIBdecl}{\relax}
\BIBdecl

\bibitem{auer2007dbpedia}
S.~Auer, C.~Bizer, G.~Kobilarov, J.~Lehmann, R.~Cyganiak, and Z.~Ives, ``Dbpedia: A nucleus for a web of open data,'' in \emph{The semantic web}.\hskip 1em plus 0.5em minus 0.4em\relax Springer, 2007, pp. 722--735.

\bibitem{suchanek2007yago}
F.~M. Suchanek, G.~Kasneci, and G.~Weikum, ``Yago: a core of semantic knowledge,'' in \emph{Proceedings of the 16th international conference on World Wide Web}, 2007, pp. 697--706.

\bibitem{zhang2020hkgb}
Y.~Zhang, M.~Sheng, R.~Zhou, Y.~Wang, G.~Han, H.~Zhang, C.~Xing, and J.~Dong, ``Hkgb: an inclusive, extensible, intelligent, semi-auto-constructed knowledge graph framework for healthcare with clinicians’ expertise incorporated,'' \emph{Information Processing \& Management}, vol.~57, no.~6, p. 102324, 2020.

\bibitem{wang2019kgat}
X.~Wang, X.~He, Y.~Cao, M.~Liu, and T.-S. Chua, ``Kgat: Knowledge graph attention network for recommendation,'' in \emph{Proceedings of the 25th ACM SIGKDD international conference on knowledge discovery \& data mining}, 2019, pp. 950--958.

\bibitem{abu2021relational}
B.~Abu-Salih, M.~Al-Tawil, I.~Aljarah, H.~Faris, P.~Wongthongtham, K.~Y. Chan, and A.~Beheshti, ``Relational learning analysis of social politics using knowledge graph embedding,'' \emph{Data Mining and Knowledge Discovery}, vol.~35, no.~4, pp. 1497--1536, 2021.

\bibitem{shen2022comprehensive}
T.~Shen, F.~Zhang, and J.~Cheng, ``A comprehensive overview of knowledge graph completion,'' \emph{Knowledge-Based Systems}, p. 109597, 2022.

\bibitem{chen2020review}
X.~Chen, S.~Jia, and Y.~Xiang, ``A review: Knowledge reasoning over knowledge graph,'' \emph{Expert Systems with Applications}, vol. 141, p. 112948, 2020.

\bibitem{dai2020survey}
Y.~Dai, S.~Wang, N.~N. Xiong, and W.~Guo, ``A survey on knowledge graph embedding: Approaches, applications and benchmarks,'' \emph{Electronics}, vol.~9, no.~5, p. 750, 2020.

\bibitem{yang2017differentiable}
F.~Yang, Z.~Yang, and W.~W. Cohen, ``Differentiable learning of logical rules for knowledge base reasoning,'' \emph{Advances in neural information processing systems}, vol.~30, 2017.

\bibitem{pujara2017sparsity}
J.~Pujara, E.~Augustine, and L.~Getoor, ``Sparsity and noise: Where knowledge graph embeddings fall short,'' in \emph{Proceedings of the 2017 conference on empirical methods in natural language processing}, 2017, pp. 1751--1756.

\bibitem{yehudai2021local}
G.~Yehudai, E.~Fetaya, E.~Meirom, G.~Chechik, and H.~Maron, ``From local structures to size generalization in graph neural networks,'' in \emph{International Conference on Machine Learning}.\hskip 1em plus 0.5em minus 0.4em\relax PMLR, 2021, pp. 11\,975--11\,986.

\bibitem{bevilacqua2021size}
B.~Bevilacqua, Y.~Zhou, and B.~Ribeiro, ``Size-invariant graph representations for graph classification extrapolations,'' in \emph{International Conference on Machine Learning}.\hskip 1em plus 0.5em minus 0.4em\relax PMLR, 2021, pp. 837--851.

\bibitem{cheng2024logical}
K.~Cheng and Y.~Sun, ``Logical rule learning,'' \emph{Knowledge Graph Reasoning: A Neuro-Symbolic Perspective}, pp. 107--147, 2024.

\bibitem{li2022ood}
H.~Li, X.~Wang, Z.~Zhang, and W.~Zhu, ``Ood-gnn: Out-of-distribution generalized graph neural network,'' \emph{IEEE Transactions on Knowledge and Data Engineering}, 2022.

\bibitem{qu2020rnnlogic}
M.~Qu, J.~Chen, L.-P. Xhonneux, Y.~Bengio, and J.~Tang, ``Rnnlogic: Learning logic rules for reasoning on knowledge graphs,'' in \emph{International Conference on Learning Representations}, 2020.

\bibitem{levine2020offline}
S.~Levine, A.~Kumar, G.~Tucker, and J.~Fu, ``Offline reinforcement learning: Tutorial, review, and perspectives on open problems,'' \emph{arXiv preprint arXiv:2005.01643}, 2020.

\bibitem{hendrycks2020pretrained}
D.~Hendrycks, X.~Liu, E.~Wallace, A.~Dziedzic, R.~Krishnan, and D.~Song, ``Pretrained transformers improve out-of-distribution robustness,'' \emph{arXiv preprint arXiv:2004.06100}, 2020.

\bibitem{zhang2021deep}
X.~Zhang, P.~Cui, R.~Xu, L.~Zhou, Y.~He, and Z.~Shen, ``Deep stable learning for out-of-distribution generalization,'' in \emph{Proceedings of the IEEE/CVF Conference on Computer Vision and Pattern Recognition}, 2021, pp. 5372--5382.

\bibitem{cheng2022rlogic}
K.~Cheng, J.~Liu, W.~Wang, and Y.~Sun, ``Rlogic: Recursive logical rule learning from knowledge graphs,'' in \emph{Proceedings of the 28th ACM SIGKDD Conference on Knowledge Discovery and Data Mining}, 2022, pp. 179--189.

\bibitem{tishby2022mean}
I.~Tishby, O.~Biham, R.~K{\"u}hn, and E.~Katzav, ``The mean and variance of the distribution of shortest path lengths of random regular graphs,'' \emph{Journal of Physics A: Mathematical and Theoretical}, vol.~55, no.~26, p. 265005, 2022.

\bibitem{bordes2013transe}
A.~Bordes, N.~Usunier, A.~Garcia-Duran, J.~Weston, and O.~Yakhnenko, ``Translating embeddings for modeling multi-relational data,'' \emph{Advances in neural information processing systems}, vol.~26, 2013.

\bibitem{wang2014knowledge}
Z.~Wang, J.~Zhang, J.~Feng, and Z.~Chen, ``Knowledge graph embedding by translating on hyperplanes,'' in \emph{Proceedings of the AAAI conference on artificial intelligence}, vol.~28, no.~1, 2014.

\bibitem{lin2015learning}
Y.~Lin, Z.~Liu, M.~Sun, Y.~Liu, and X.~Zhu, ``Learning entity and relation embeddings for knowledge graph completion,'' in \emph{Proceedings of the AAAI conference on artificial intelligence}, vol.~29, no.~1, 2015.

\bibitem{sun2019rotate}
Z.~Sun, Z.-H. Deng, J.-Y. Nie, and J.~Tang, ``Rotate: Knowledge graph embedding by relational rotation in complex space,'' \emph{arXiv preprint arXiv:1902.10197}, 2019.

\bibitem{nickel2011three}
M.~Nickel, V.~Tresp, H.-P. Kriegel \emph{et~al.}, ``A three-way model for collective learning on multi-relational data.'' in \emph{Icml}, vol.~11, no. 10.5555, 2011, pp. 3\,104\,482--3\,104\,584.

\bibitem{yang2014embedding}
B.~Yang, W.-t. Yih, X.~He, J.~Gao, and L.~Deng, ``Embedding entities and relations for learning and inference in knowledge bases,'' \emph{arXiv preprint arXiv:1412.6575}, 2014.

\bibitem{trouillon2016complex}
T.~Trouillon, J.~Welbl, S.~Riedel, {\'E}.~Gaussier, and G.~Bouchard, ``Complex embeddings for simple link prediction,'' in \emph{International conference on machine learning}.\hskip 1em plus 0.5em minus 0.4em\relax PMLR, 2016, pp. 2071--2080.

\bibitem{schlichtkrull2018modeling}
M.~Schlichtkrull, T.~N. Kipf, P.~Bloem, R.~Van Den~Berg, I.~Titov, and M.~Welling, ``Modeling relational data with graph convolutional networks,'' in \emph{The semantic web: 15th international conference, ESWC 2018, Heraklion, Crete, Greece, June 3--7, 2018, proceedings 15}.\hskip 1em plus 0.5em minus 0.4em\relax Springer, 2018, pp. 593--607.

\bibitem{wang2019robust}
Z.~Wang, Z.~Ren, C.~He, P.~Zhang, and Y.~Hu, ``Robust embedding with multi-level structures for link prediction.'' in \emph{IJCAI}, 2019, pp. 5240--5246.

\bibitem{vashishthRGCN}
S.~Vashishth, S.~Sanyal, V.~Nitin, and P.~Talukdar, ``Composition-based multi-relational graph convolutional networks,'' in \emph{International Conference on Learning Representations}.

\bibitem{teru2020inductive}
K.~Teru, E.~Denis, and W.~Hamilton, ``Inductive relation prediction by subgraph reasoning,'' in \emph{International Conference on Machine Learning}.\hskip 1em plus 0.5em minus 0.4em\relax PMLR, 2020, pp. 9448--9457.

\bibitem{zhu2021neural}
Z.~Zhu, Z.~Zhang, L.-P. Xhonneux, and J.~Tang, ``Neural bellman-ford networks: A general graph neural network framework for link prediction,'' \emph{Advances in neural information processing systems}, vol.~34, pp. 29\,476--29\,490, 2021.

\bibitem{zhu2023net}
Z.~Zhu, X.~Yuan, M.~Galkin, L.-P. Xhonneux, M.~Zhang, M.~Gazeau, and J.~Tang, ``A* net: A scalable path-based reasoning approach for knowledge graphs,'' \emph{Advances in Neural Information Processing Systems}, vol.~36, pp. 59\,323--59\,336, 2023.

\bibitem{zhang2022knowledge}
Y.~Zhang and Q.~Yao, ``Knowledge graph reasoning with relational digraph,'' in \emph{Proceedings of the ACM web conference 2022}, 2022, pp. 912--924.

\bibitem{muggleton1994inductive}
S.~Muggleton and L.~De~Raedt, ``Inductive logic programming: Theory and methods,'' \emph{The Journal of Logic Programming}, vol.~19, pp. 629--679, 1994.

\bibitem{muggleton1990efficient}
S.~H. Muggleton, C.~Feng \emph{et~al.}, \emph{Efficient induction of logic programs}.\hskip 1em plus 0.5em minus 0.4em\relax Turing Institute, 1990.

\bibitem{rocktaschel2017end}
T.~Rockt{\"a}schel and S.~Riedel, ``End-to-end differentiable proving,'' \emph{Advances in neural information processing systems}, vol.~30, 2017.

\bibitem{sadeghian2019drum}
A.~Sadeghian, M.~Armandpour, P.~Ding, and D.~Z. Wang, ``Drum: End-to-end differentiable rule mining on knowledge graphs,'' \emph{Advances in Neural Information Processing Systems}, vol.~32, pp. 15\,347--15\,357, 2019.

\bibitem{cheng2022neural}
K.~Cheng, N.~Ahmed, and Y.~Sun, ``Neural compositional rule learning for knowledge graph reasoning,'' in \emph{The Eleventh International Conference on Learning Representations}, 2022.

\bibitem{li2018learning}
D.~Li, Y.~Yang, Y.-Z. Song, and T.~Hospedales, ``Learning to generalize: Meta-learning for domain generalization,'' in \emph{Proceedings of the AAAI conference on artificial intelligence}, vol.~32, no.~1, 2018.

\bibitem{li2018domain}
H.~Li, S.~J. Pan, S.~Wang, and A.~C. Kot, ``Domain generalization with adversarial feature learning,'' in \emph{Proceedings of the IEEE conference on computer vision and pattern recognition}, 2018, pp. 5400--5409.

\bibitem{dou2019domain}
Q.~Dou, D.~Coelho~de Castro, K.~Kamnitsas, and B.~Glocker, ``Domain generalization via model-agnostic learning of semantic features,'' \emph{Advances in neural information processing systems}, vol.~32, 2019.

\bibitem{hu2020domain}
S.~Hu, K.~Zhang, Z.~Chen, and L.~Chan, ``Domain generalization via multidomain discriminant analysis,'' in \emph{Uncertainty in Artificial Intelligence}.\hskip 1em plus 0.5em minus 0.4em\relax PMLR, 2020, pp. 292--302.

\bibitem{piratla2020efficient}
V.~Piratla, P.~Netrapalli, and S.~Sarawagi, ``Efficient domain generalization via common-specific low-rank decomposition,'' in \emph{International Conference on Machine Learning}.\hskip 1em plus 0.5em minus 0.4em\relax PMLR, 2020, pp. 7728--7738.

\bibitem{seo2020learning}
S.~Seo, Y.~Suh, D.~Kim, G.~Kim, J.~Han, and B.~Han, ``Learning to optimize domain specific normalization for domain generalization,'' in \emph{Computer Vision--ECCV 2020: 16th European Conference, Glasgow, UK, August 23--28, 2020, Proceedings, Part XXII 16}.\hskip 1em plus 0.5em minus 0.4em\relax Springer, 2020, pp. 68--83.

\bibitem{carlucci2019domain}
F.~M. Carlucci, A.~D'Innocente, S.~Bucci, B.~Caputo, and T.~Tommasi, ``Domain generalization by solving jigsaw puzzles,'' in \emph{Proceedings of the IEEE/CVF Conference on Computer Vision and Pattern Recognition}, 2019, pp. 2229--2238.

\bibitem{shankar2018generalizing}
S.~Shankar, V.~Piratla, S.~Chakrabarti, S.~Chaudhuri, P.~Jyothi, and S.~Sarawagi, ``Generalizing across domains via cross-gradient training,'' \emph{arXiv preprint arXiv:1804.10745}, 2018.

\bibitem{volpi2018generalizing}
R.~Volpi, H.~Namkoong, O.~Sener, J.~C. Duchi, V.~Murino, and S.~Savarese, ``Generalizing to unseen domains via adversarial data augmentation,'' \emph{Advances in neural information processing systems}, vol.~31, 2018.

\bibitem{li2019episodic}
D.~Li, J.~Zhang, Y.~Yang, C.~Liu, Y.-Z. Song, and T.~M. Hospedales, ``Episodic training for domain generalization,'' in \emph{Proceedings of the IEEE/CVF International Conference on Computer Vision}, 2019, pp. 1446--1455.

\bibitem{arjovsky2019invariant}
M.~Arjovsky, L.~Bottou, I.~Gulrajani, and D.~Lopez-Paz, ``Invariant risk minimization,'' \emph{arXiv preprint arXiv:1907.02893}, 2019.

\bibitem{kuang2020stable}
K.~Kuang, R.~Xiong, P.~Cui, S.~Athey, and B.~Li, ``Stable prediction with model misspecification and agnostic distribution shift,'' in \emph{Proceedings of the AAAI Conference on Artificial Intelligence}, vol.~34, no.~04, 2020, pp. 4485--4492.

\bibitem{shen2020stable}
Z.~Shen, P.~Cui, T.~Zhang, and K.~Kunag, ``Stable learning via sample reweighting,'' in \emph{Proceedings of the AAAI Conference on Artificial Intelligence}, vol.~34, no.~04, 2020, pp. 5692--5699.

\bibitem{xu2021stable}
R.~Xu, P.~Cui, Z.~Shen, X.~Zhang, and T.~Zhang, ``Why stable learning works? a theory of covariate shift generalization,'' \emph{arXiv preprint arXiv:2111.02355}, vol.~2, 2021.

\bibitem{dou2022decorrelate}
S.~Dou, R.~Zheng, T.~Wu, S.~Gao, J.~Shan, Q.~Zhang, Y.~Wu, and X.~Huang, ``Decorrelate irrelevant, purify relevant: Overcome textual spurious correlations from a feature perspective,'' \emph{arXiv preprint arXiv:2202.08048}, 2022.

\bibitem{salvat1996sound}
E.~Salvat and M.-L. Mugnier, ``Sound and complete forward and backward chainings of graph rules,'' in \emph{International Conference on Conceptual Structures}.\hskip 1em plus 0.5em minus 0.4em\relax Springer, 1996, pp. 248--262.

\bibitem{galarraga2013amie}
L.~A. Gal{\'a}rraga, C.~Teflioudi, K.~Hose, and F.~Suchanek, ``Amie: association rule mining under incomplete evidence in ontological knowledge bases,'' in \emph{Proceedings of the 22nd international conference on World Wide Web}, 2013, pp. 413--422.

\bibitem{chen2016robust}
X.~Chen, M.~Monfort, A.~Liu, and B.~D. Ziebart, ``Robust covariate shift regression,'' in \emph{Artificial Intelligence and Statistics}.\hskip 1em plus 0.5em minus 0.4em\relax PMLR, 2016, pp. 1270--1279.

\bibitem{sugiyama2013learning}
M.~Sugiyama, M.~Yamada, and M.~C. du~Plessis, ``Learning under nonstationarity: covariate shift and class-balance change,'' \emph{Wiley Interdisciplinary Reviews: Computational Statistics}, vol.~5, no.~6, pp. 465--477, 2013.

\bibitem{xu2022theoretical}
R.~Xu, X.~Zhang, Z.~Shen, T.~Zhang, and P.~Cui, ``A theoretical analysis on independence-driven importance weighting for covariate-shift generalization,'' in \emph{International Conference on Machine Learning}.\hskip 1em plus 0.5em minus 0.4em\relax PMLR, 2022, pp. 24\,803--24\,829.

\bibitem{tu2020empirical}
L.~Tu, G.~Lalwani, S.~Gella, and H.~He, ``An empirical study on robustness to spurious correlations using pre-trained language models,'' \emph{Transactions of the Association for Computational Linguistics}, vol.~8, pp. 621--633, 2020.

\bibitem{bisgaard2006does}
T.~M. Bisgaard and Z.~Sasv{\'a}ri, ``When does e (xk{\textperiodcentered} yl)= e (xk){\textperiodcentered} e (yl) imply independence?'' \emph{Statistics \& probability letters}, vol.~76, no.~11, pp. 1111--1116, 2006.

\bibitem{athey2018approximate}
S.~Athey, G.~W. Imbens, and S.~Wager, ``Approximate residual balancing: debiased inference of average treatment effects in high dimensions,'' \emph{Journal of the Royal Statistical Society Series B: Statistical Methodology}, vol.~80, no.~4, pp. 597--623, 2018.

\bibitem{fong2018covariate}
C.~Fong, C.~Hazlett, and K.~Imai, ``Covariate balancing propensity score for a continuous treatment: Application to the efficacy of political advertisements,'' \emph{The Annals of Applied Statistics}, vol.~12, no.~1, pp. 156--177, 2018.

\bibitem{hollander2013nonparametric}
M.~Hollander, D.~A. Wolfe, and E.~Chicken, \emph{Nonparametric statistical methods}.\hskip 1em plus 0.5em minus 0.4em\relax John Wiley \& Sons, 2013.

\bibitem{spitzer2013principles}
F.~Spitzer, \emph{Principles of random walk}.\hskip 1em plus 0.5em minus 0.4em\relax Springer Science \& Business Media, 2013, vol.~34.

\bibitem{hinton1986learning}
G.~E. Hinton \emph{et~al.}, ``Learning distributed representations of concepts,'' in \emph{Proceedings of the eighth annual conference of the cognitive science society}, vol.~1.\hskip 1em plus 0.5em minus 0.4em\relax Amherst, MA, 1986, p.~12.

\bibitem{kok2007statistical}
S.~Kok and P.~Domingos, ``Statistical predicate invention,'' in \emph{Proceedings of the 24th international conference on Machine learning}, 2007, pp. 433--440.

\bibitem{dettmers2018convolutional}
T.~Dettmers, P.~Minervini, P.~Stenetorp, and S.~Riedel, ``Convolutional 2d knowledge graph embeddings,'' in \emph{Proceedings of the AAAI conference on artificial intelligence}, vol.~32, no.~1, 2018.

\bibitem{toutanova2015observed}
K.~Toutanova and D.~Chen, ``Observed versus latent features for knowledge base and text inference,'' in \emph{Proceedings of the 3rd workshop on continuous vector space models and their compositionality}, 2015, pp. 57--66.

\bibitem{tang2008arnetminer}
J.~Tang, J.~Zhang, L.~Yao, J.~Li, L.~Zhang, and Z.~Su, ``Arnetminer: extraction and mining of academic social networks,'' in \emph{Proceedings of the 14th ACM SIGKDD international conference on Knowledge discovery and data mining}, 2008, pp. 990--998.

\bibitem{chen2013geotext}
X.~Chen, D.~Wang, and T.~Zhao, ``Geotext: an intelligent dynamic geometry textbook,'' \emph{ACM Communications in Computer Algebra}, vol.~46, no. 3/4, pp. 171--175, 2013.

\bibitem{paszke2019pytorch}
A.~Paszke, S.~Gross, F.~Massa, A.~Lerer, J.~Bradbury, G.~Chanan, T.~Killeen, Z.~Lin, N.~Gimelshein, L.~Antiga \emph{et~al.}, ``Pytorch: An imperative style, high-performance deep learning library,'' \emph{Advances in neural information processing systems}, vol.~32, 2019.

\bibitem{kingma2014adam}
D.~P. Kingma and J.~Ba, ``Adam: A method for stochastic optimization,'' \emph{arXiv preprint arXiv:1412.6980}, 2014.

\bibitem{sun2022heterogeneous}
Y.~Sun, J.~Han, X.~Yan, P.~S. Yu, and T.~Wu, ``Heterogeneous information networks: the past, the present, and the future,'' \emph{Proceedings of the VLDB Endowment}, vol.~15, no.~12, pp. 3807--3811, 2022.

\end{thebibliography}
